\documentclass[nohyperref]{article}

\usepackage{microtype}
\usepackage{graphicx}
\usepackage{subfigure}
\usepackage{booktabs} 

\usepackage{hyperref}

\usepackage[accepted]{icml2022}

\usepackage{amsmath}
\usepackage{amssymb}
\usepackage{mathtools}
\usepackage{amsthm}

\usepackage{times}
\usepackage{soul}
\usepackage{url}
\usepackage{comment}
\usepackage[utf8]{inputenc}
\usepackage{graphicx}
\usepackage{amsmath}
\usepackage{amsthm}
\usepackage{booktabs}
\usepackage{algorithm}
\usepackage{algorithmic}

\usepackage{amssymb}
\usepackage{bbm}
\usepackage{array}
\usepackage{multicol}
\usepackage{multirow}

\usepackage[capitalize,noabbrev]{cleveref}

\theoremstyle{plain}

\theoremstyle{definition}

\theoremstyle{remark}

\usepackage[textsize=tiny]{todonotes}

\icmltitlerunning{Gaussian Mixture VAE with Contrastive Learning for Multi-Label Classification}

\begin{document}

\twocolumn[
\icmltitle{Gaussian Mixture Variational Autoencoder with Contrastive Learning \\ for Multi-Label Classification}



\icmlsetsymbol{equal}{*}

\begin{icmlauthorlist}
\icmlauthor{Junwen Bai}{co}
\icmlauthor{Shufeng Kong}{co}
\icmlauthor{Carla Gomes}{co}
\end{icmlauthorlist}

\icmlaffiliation{co}{Department of Computer Science, Cornell University, Ithaca, USA}

\icmlcorrespondingauthor{Shufeng Kong}{sk2299@cornell.edu}

\icmlkeywords{contrastive learning, VAE, Gaussian mixture, multi-label classification}

\vskip 0.3in
]



\printAffiliationsAndNotice{}  

\begin{abstract}
Multi-label classification (MLC) is a prediction task where each sample can have more than one label. We propose a novel contrastive learning boosted multi-label prediction model based on a Gaussian mixture variational autoencoder (C-GMVAE), which learns a multimodal prior space and employs a contrastive loss. Many existing methods introduce extra complex neural modules like graph neural networks to capture the label correlations, in addition to the prediction modules. We find that by using contrastive learning in the supervised setting, we can exploit label information effectively in a data-driven manner, and learn meaningful feature and label embeddings which capture the label correlations and enhance the predictive power. Our method also adopts the idea of learning and aligning latent spaces for both features and labels. In contrast to previous works based on a unimodal prior, C-GMVAE imposes a Gaussian mixture structure on the latent space, to alleviate the posterior collapse and over-regularization issues. C-GMVAE outperforms existing methods on multiple public datasets and can often match other models' full performance with only 50\% of the training data. Furthermore, we show that the learnt embeddings provide insights into the interpretation of label-label interactions.
\end{abstract}

\section{Introduction}

In many machine learning tasks, an instance can have several labels. The task of predicting multiple labels is known as multi-label classification (MLC). MLC is
common
in domains like computer vision \cite{wang2016cnn}, natural language processing \cite{chang2019x} and biology \cite{yu2013protein}.  Unlike the single-label scenario, label correlations are more important in MLC. Early works capture the correlations through classifier chains \cite{read2009classifier}, Bayesian inference \cite{zhang2007ml}, and dimensionality reduction \cite{bhatia2015sparse}.

Thanks to the huge capacity of neural networks (NN), many previous methods can be improved by their neural extensions. For example, classifier chains can be naturally enhanced by recurrent neural networks (RNN) \cite{wang2016cnn}. The non-linearity of NN alleviates the complex design of feature mapping and many deep models can therefore focus on the loss function, feature-label and label-label correlation modeling. 

One trending direction is to learn a deep latent space shared by features and labels. The encoded samples from the latent space are then decoded to targets. One typical example is C2AE \cite{yeh2017learning}, which learns latent codes for both features and labels. The latent codes are passed to a decoder to derive the target labels. C2AE minimizes an $\ell_2$ distance between the feature and label codes, together with a relaxed orthogonality regularization. However, the learnt deterministic latent space lacks smoothness and structures. Small perturbations in this latent space can lead to totally different decoding results. Even if the corresponding feature and label codes are close, we cannot guarantee the decoded targets are similar. To address this concern, MPVAE \cite{bai2020disentangled} proposes to replace the deterministic latent space with a probabilistic space under a variational autoencoder (VAE) framework. The Gaussian latent spaces are aligned with KL-divergence, and the sampling process enforces smoothness. Similar ideas can be found in \cite{sundar2020out}. However, these methods assume a unimodal Gaussian latent space, which is known to cause over-regularization and posterior collapse \cite{dilokthanakul2016deep,wu2018multimodal}. A better strategy would be to learn a multimodal latent space. It is more reasonable to assume the observed data are generated from a multimodal subspace rather than a unimodal one. 

Another popular group of methods focuses on better label correlation modeling. Their idea is straightforward: some labels should be more correlated if they co-appear often while others should be less relevant. Existing methods adopt pairwise ranking loss, covariance matrices, conditional random fields (CRF) or graph neural nets (GNN) to this end \cite{zhang2013review,bi2014multilabel,belanger2016structured,lanchantin2019neural,chen2019multi}. These methods often either constrain the learning through a predefined structure (which requires a larger model size), or aren't powerful enough to capture the correlations (such as pairwise ranking loss). 

Our idea is simple: we learn embeddings for each label class and the inner products between embeddings should reflect the similarity. We further learn feature embeddings whose inner products with label embeddings correspond to feature-label similarity and can be used for prediction. We assume these embeddings are generated from a probabilistic multimodal latent space shared by features and labels, where we use KL-divergence to align the feature and label latent distributions. On the other hand, one might be concerned that embeddings alone won't capture both label-label and label-feature correlations, which were usually modeled by extra GNN and covariance matrices in prior works \cite{lanchantin2019neural,bai2020disentangled}. To this end, we stress on the loss function terms rather than extra structure to capture these correlations. Intuitively, if two labels co-appear often, their embeddings should be close. Otherwise, if two labels seldom co-appear, their embeddings should be distant. A triplet-like loss could be naturally applied in this scenario. Nonetheless, its extension, contrastive loss, has shown to be even more effective than the triplet loss by introducing more samples rather than just one triplet. We show that contrastive loss can pull together correlated label embeddings, push away unrelated label embeddings (see Fig.~\ref{fig:bird}), and even perform better than GNN-based or covariance-based methods.


Our new model for MLC, contrastive learning boosted Gaussian mixture variational autoencoder (C-GMVAE), alleviates the over-regularization and posterior collapse concerns, and also learns useful feature and label embeddings. C-GMVAE is applied to nine datasets and outperforms the existing methods on five metrics. Moreover, we show that using only 50\% of the data, our results can match the full performance of other state-of-the-art methods. Ablation studies and interpretability of learnt embeddings will also be illustrated in the experiments. Our contributions can be summarized in three aspects: \textbf{(i)} 
We propose to use contrastive loss instead of triplet or ranking loss to strengthen the label embedding learning.
We empirically show that by using a contrastive loss, one can get rid of heavy-duty label correlation modules (e.g., covariance matrices, GNNs) while achieving even better performances. \textbf{(ii)} Though contrastive learning is commonly applied in self-supervised learning, our work shows that by properly defining anchor, positive and negative samples, contrastive loss can leverage label information very effectively in the supervised MLC scenario as well. \textbf{(iii)} Unlike prior probabilistic models, C-GMVAE learns a multimodal latent space and integrates the probabilistic modeling (VAE module) with embedding learning (contrastive module) synergistically. 

\section{Methods}

In MLC, given a dataset containing $N$ labeled samples $(x,y)$, where $x\in \mathbb R^D$ and $y\in \mathbb \{0,1\}^L$, our goal is to find a mapping from $x$ to $y$. $N, D, L$ are the number of samples, feature length and label set size respectively. The binary coding indicates the labels associated with the sample $x$. Labels are correlated with each other.

\subsection{Preliminaries}

\subsubsection{Gaussian Mixture VAE}
A standard VAE \cite{kingma2013auto} pulls together the posterior distribution and a parameter-free isotropic Gaussian prior. Two losses are optimized together in training: KL-divergence from the prior to the posterior, and the distance between the reconstructed targets and the real targets. One weakness of this formulation is the unimodality of its latent space, inhibiting the learning of more complex representations. Another concern is over-regularization: if the posterior is exactly the same as the prior, the learnt representations would be uninformative of the inputs.
Numerous works extend the prior to be more complex \cite{chung2015recurrent,eslami2016attend,dilokthanakul2016deep}. In our work, we adopt the Gausian mixture prior. The probability density can be depicted as $p(z)=\frac{1}{k}\sum_{i=1}^k\mathcal N(z|\mu_i,\sigma_i^2)$ where $i$ is the cluster index of $k$ Gaussian clusters with mean $\mu_i$ and covariance $\sigma_i^2$ \cite{shu2016gaussian, shi2019variational}. Our intuition is that each label embedding could correlate to a Gaussian subspace. Given a label set, the mixture of the positive Gaussian subspaces forms a unique multimodal prior distribution. The label embeddings also receive the gradients from the contrastive loss and thus the contrastive learning is combined with latent space construction. Our formulation is also related to MVAE \cite{wu2018multimodal,shi2020relating} 
which adopts the idea of product-of-experts.

\begin{figure*}[t]
    \centering
    \includegraphics[width=0.99\textwidth]{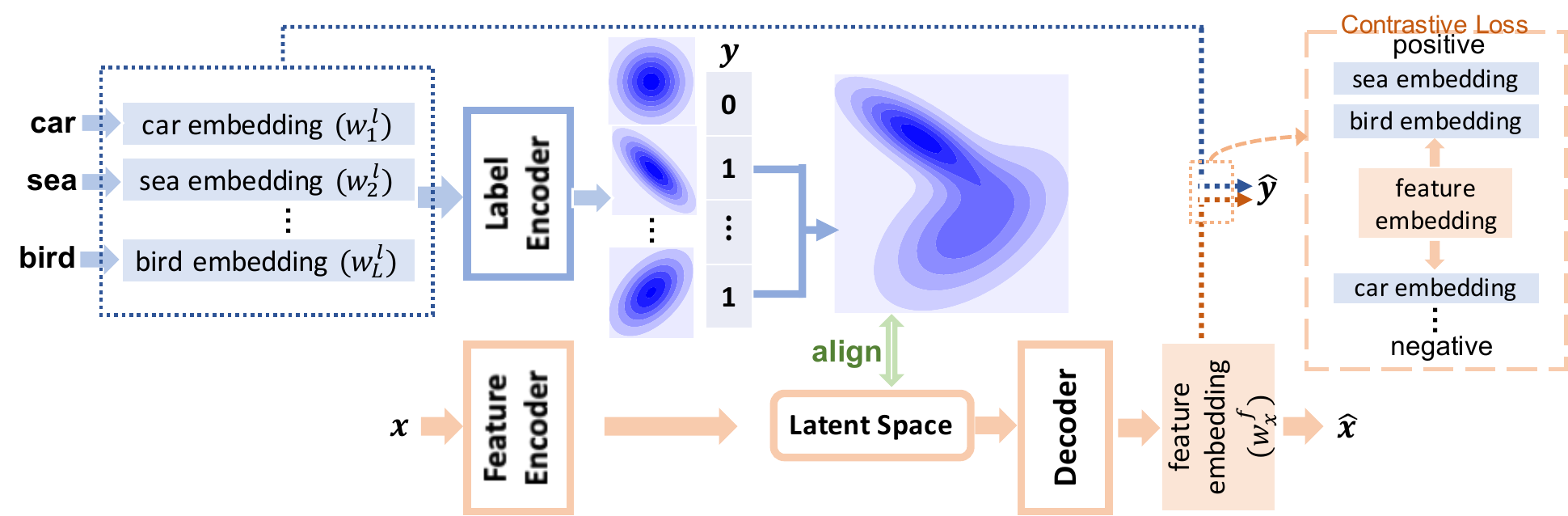}
    \caption{The full pipeline of C-GMVAE. Every label category is mapped to a learnable embedding first. The label encoder transforms each embedding $w_i^l$ to a multivariate Gaussian latent space. The sample's associated label set selects the positive latent spaces and forms a Gaussian mixture prior. Each feature is also mapped to a latent space through a feature encoder. The posterior is aligned with the prior via KL-divergence. The decoder takes in a sample from the latent space and produces a feature embedding $w_x^f$. A contrastive loss is designed to pull together the feature embedding and the positive label embeddings, while separating the feature embedding from the negative label embeddings. Prediction $\hat{y}$ is generated by passing the inner products between the feature embedding $w_x^f$ and the label embeddings $w_i^l$ to the sigmoid functions. A sample with label set \{sea, bird\} is shown here.}
    \label{fig:cgmvae}
\end{figure*}

\subsubsection{Contrastive Learning}
We propose to use contrastive learning to capture the correlations (i.e., feature-label and label-label correlations).
Contrastive learning \cite{oord2018representation, chen2020simple,khosla2020supervised} is a novel learning style. The core idea is simple: given an anchor sample, it should be close to similar samples (positive) and far from dissimilar samples (negative) in some learnt embedding space. It differs from triplet loss in the number of negative samples and the loss estimation method. 
Contrastive loss is largely motivated by noise contrastive estimation (NCE) \cite{gutmann2010noise} and its form is generalizable. 
The raw contrastive loss formulation only considers the instance-level invariance (multiple views of one instance), but with label information, we can learn category-level invariance (multiple instances per class/category) \cite{wang2020unsupervised}. 

In the multi-label scenario, one can regard the feature embedding as the anchor sample, positive label embeddings as the positive samples and negative label embeddings as the negative samples. The formulation can fit the contrastive learning framework naturally and is one of our major contributions. Compared to the pairwise ranking loss which focuses on the final logits, contrastive loss is defined on the learnt embeddings and thus becomes more expressive. Contrastive loss also includes more samples in estimating the NCE and therefore outperforms the triplet loss. In the appendix, we show triplet loss is actually a special case of our contrastive loss. 

\subsection{C-GMVAE} \label{c-gmvae}

C-GMVAE inherits the general VAE framework, but with a learnable Gaussian mixture (GM) prior. 
During training, each sample's label set activates and mixes the positive Gaussian subspaces to derive the prior. 
Contrastive learning is applied to boost the embedding learning, using a contrastive loss between the feature and label embeddings. Fig.~\ref{fig:cgmvae} provides a full illustration, and the following subsections will elaborate on the details.

\subsubsection{Gaussian Mixture Latent Space}

Given a sample $(x, y)$ where feature $x\in \mathbb R^D$ and label $y\in \{0,1\}^L$, many previous works take $y$ as the input and transform it to a dense representation through a fully-connected layer~\cite{yeh2017learning,bai2020disentangled}. 
This layer essentially maps each label category to an embedding and sums up all the embeddings using label $y$ as weights (0 or 1). The summed embedding is fed into the label encoder to produce a probabilistic latent space. 

In C-GMVAE, however, we directly map each label embedding $w_i^l\in\mathbb R^E$ of label class $i$ to an individual latent Gaussian distribution $\mathcal N(\mu_i,diag(\sigma_i^2))$, where $\mu_i\in\mathbb R^d, \sigma_i^2\in\mathbb R^d$, and  $\mu_i,\sigma_i^2$ are derived from $w_i^l$ through the NN-based label encoder. The randomly initialized embeddings $w_i^l$ are learnable during the training process, similar to \cite{mikolov2013efficient}, and they share the same label encoder. In Fig.~\ref{fig:cgmvae}, the label categories \textit{car, sea,..., bird} are transformed to embeddings first. Embeddings are then passed directly to label encoder rather than summed up. Each label category (e.g., \textit{car}) corresponds to a unimodal Gaussian in the latent space. $y$ activates ``positive" Gaussians ($y_i=1$) and forms a Gaussian mixture subspace.
Given a random variable $z\in\mathbb R^d$, the probability density function (PDF) in the subspace is defined as
\begin{equation}
\begin{aligned}
p_\psi(z|y)=\frac{1}{\sum_i y_i}\sum_{i=1}^L\mathbbm 1\{y_i=1\}\mathcal N(z|\mu_i,diag(\sigma_i^2))
\end{aligned}
\end{equation}
where $\mathbbm 1(\cdot)$ is the indicator function and the label encoder is parameterized by $\psi$ (NN). In Fig.~\ref{fig:cgmvae}, $y$ activates \textit{sea} and \textit{bird}, then we have 
\begin{equation}
\begin{aligned}p_\psi(z|y)=\frac{1}{2}(&\mathcal N(z|\mu_{sea},diag(\sigma_{sea}^2))+\\
&\mathcal N(z|\mu_{bird},diag(\sigma_{bird}^2)))
\end{aligned}
\end{equation}

Most VAE-based frameworks optimize over an evidence lower bound (ELBO) \cite{doersch2016tutorial}:
\begin{equation}
\begin{aligned}
\text{ELBO}=\mathbb E_{q_\phi(z|x)}[&\log p_\theta(x|z)]-\\
&D_{KL}[q_\phi(z|x)||p(z)]
\end{aligned}
\end{equation}
The feature encoder is parameterized by $\phi$ (NN). One pitfall of this objective is owing to the minimization of the KL-divergence. If the divergence between the posterior $q_\phi(z|x)$ and the prior $p_\psi(z)$ vanishes, the learnt latent codes would become non-informative. This is called posterior collapse. Many recent works suggest learnable priors \cite{tomczak2018vae} or more sophisticated priors \cite{wang2019neural} to avoid this issue, and we adopt these ideas in our design of the prior. Compared to a standard VAE, our prior is informative, learnable and multimodal.

We form a standard posterior in our model and match it with the prior. 
However, unlike vanilla VAE, we cannot analytically compute the KL term. Instead, we use the following estimation:
\begin{equation}
\begin{aligned}
\mathcal L_{KL}\approx& \log q_\phi(z_0|x)-\log p_\psi(z_0|y)\\
=&\log \mathcal N(z_0|\mu_\phi(x),diag(\sigma_\phi^2(x)))-\\
&\log \frac{1}{\sum_i y_i}\sum_{i=1}^L\mathbbm 1\{y_i=1\}\mathcal N(z_0|\mu_i,diag(\sigma_i^2))
\end{aligned}
\end{equation}
where $z_0\sim q_\phi(z|x)$ denotes a single latent sample. Our formulation follows the design of \cite{shu2016gaussian}, which has been shown to outperform the formulation in \cite{dilokthanakul2016deep}.

The reconstruction loss is a standard negative log-likelihood with decoder parameters $\theta$,
\begin{equation}
\begin{aligned}
\mathcal L_{recon}=& -E_{q_\phi(z|x)}[\log p_\theta(x|z)]
\end{aligned}
\end{equation}

\subsubsection{Contrastive Learning Module}

The decoder function $f_\theta^d(\cdot)$ decodes the sample from the latent space to a feature embedding $w^f_x\in \mathbb R^E$. We learn $w^f_x$ together with label embeddings $\{w_i^l\}_{i=1}^L$. The objective function includes both contrastive loss and cross-entropy loss terms.


Prior works explicitly capture the label-label interactions with GNNs or covariance matrices, which impose the structure \textit{a priori} and might not be the best modeling approach. Our contrastive module instead captures the correlations in a data-driven manner. For example, if in most of the samples, ``beach" and ``sunshine" appear together, the contrastive learning will implicitly pull their embeddings together (see the derivation in the appendix). In other words, if two labels do co-appear often, their label embeddings would become similar (Fig.~\ref{fig:bird}). On the other hand, if they never co-occur or only co-appear occasionally, their connections are not significant and our model will not optimize for their similarity. 

Original contrastive learning \cite{oord2018representation} augments inputs and learns the instance-level invariance, but it may not generalize to the category-level invariance. In the supervised setting, however, the learning can benefit from the labels and discover the category-level invariance \cite{khosla2020supervised}. Let $A\equiv\{1...L\}$. We define $P(y)\equiv\{i\in A:y_i=1\}$ for sample $(x,y)$. Suppose we have a batch of samples, $\mathcal B$, the contrastive loss can be written as
\begin{equation}
\begin{aligned}
\mathcal L_{CL}=\frac{1}{|\mathcal B|}\sum_{(x,y)\in \mathcal B}\frac{1}{|P(y)|}\sum_{p\in P(y)}-\log \frac{\text{sim}(w^f_x, w_p^l)}{\sum_{t\in A}\text{sim}(w^f_x, w_t^l)}
\end{aligned}
\label{eq:cl}
\end{equation}
Here, $\text{sim}(\cdot)$ is a function measuring the similarity between two embeddings,
and $w_x^f$, $w_i^l$ denote the feature and label embeddings respectively. 
Eq.~\ref{eq:cl} is built on top of NCE \cite{gutmann2010noise}, and the equation is equivalent to a categorical cross-entropy of correctly predicting positive labels.
The choice of $\text{sim}(\cdot)$ can be a log-bilinear function \cite{oord2018representation}, or a more complicated neural metric function \cite{chen2020simple}. In our experiments, we find it is simple and effective to take $\text{sim}(w_1,w_2)=\exp(w_1\cdot w_2/\tau)$ where $\cdot$ means inner product and $\tau$ is a temperature parameter controlling the scale of the inner product. 

In the single-label scenarios like SupCon \cite{khosla2020supervised}, if one class is positive, all other classes are contrastive to it. However, in MLC, if ``beach" is positive in the label set while ``sea" is not for one particular sample, we cannot say these two classes are contrastive. Their correlations should be captured implicitly by all the samples. Therefore, we do not enforce contrastive relations between labels and thus preserve the label correlations. 
Instead, we choose the feature embedding to be the anchor and label embeddings to be the positive and negative samples. If two label embeddings co-appear often as positive samples, they would implicitly become similar (see Fig.~\ref{fig:bird}).
Eq.~\ref{eq:cl} saves the effort of manually configuring the positive and negative samples, and is totally data-driven. The number of positive or negative samples could be greater than one, depending on the label set. Though $L$ limits the max samples we can have, this formulation has already used many more samples compared to triplet loss, and we will show in experiments that this formulation is very effective. 

The triplet loss often used in multi-label learning \cite{seymour2018multi} can be seen as a special case of Eq.~\ref{eq:cl} with only one positive and one negative. We illustrate this connection in the appendix. Furthermore, one desired property of embedding learning is that when a good positive embedding is already close enough to our anchor embedding, it contributes less to the gradients, while poorly learnt embeddings contribute more to improve the model performance. In the appendix, we also show that the contrastive loss can implicitly achieve this goal and a full derivation of the gradients is provided.

Our objective function also includes a supervised cross-entropy loss term to further facilitate the training. With the label embeddings $w_i^l$ and the feature embedding $w_x^f$, the cross entropy loss for each $(x,y)$ is given by
\begin{equation}
\begin{aligned}
\mathcal L_{CE}=\sum_{i=1}^Ly_i\log s(w_x^fw_i^l)+(1-y_i)\log (1-s(w_x^fw_i^l))
\end{aligned}
\end{equation}
where $s(\cdot)$ is the sigmoid function. 
In self-supervised learning, the contrastive loss typically helps the pretraining stage and the learnt representations are applied to downstream tasks. In the supervised setting, though some models \cite{khosla2020supervised} stick to the two-stage training process where the model is trained with contrastive loss in the first stage and with cross-entropy loss in the second stage, we did not observe its superiority over the one-stage scheme in our MLC scenario. This is partly because we also learn a latent space that is closely connected to label embeddings. 
We train the model with an objective function incorporating all the losses.  A joint training strategy reconciles different modules. We show in the experiments that the learnt embeddings are semantically meaningful and can reveal the label correlations.

\subsubsection{Objective Function}

The final objective function to minimize is simply the summation of different losses,
\begin{equation}
\mathcal L=\mathcal L_{KL}+\mathcal L_{recon}+\alpha\mathcal L_{CL}-\beta\mathcal L_{CE}
\label{eq:final_obj}
\end{equation}
where $\alpha,\beta$ are trade-off weights. The model is trained with Adam \cite{kingma2014adam}. Our model is optimized with $\mathcal L$ and will be tested on five different metrics. This is different from the methods that only optimize and test for specific metrics \cite{koyejo2015consistent,decubber2018deep}.

\subsection{Prediction}

During the testing phase, the input sample $x$ will be passed to the feature encoder and decoder to obtain its embedding $w_x^f$. Label embeddings $w_i^l$ are fixed in testing. The inner products between $w_x^f$ and $w_i^l$ will be passed through a sigmoid function to obtain the prediction probability for each class $i$.

\subsection{Insights behind C-GMVAE}
C2AE and MPVAE have shown the importance of learning a shared latent space for both features and labels. These methods share the same high-level insight similar to a teacher-student regime: we map labels (teacher) to a latent space with some certain structure, which preserves the label information and is easier to decode back to labels. Then the features (student) are expected to be mapped to this latent space to facilitate the label prediction. Two general concerns exist for these methods: 1) the unimodal Gaussian space previously used is too restrictive to impose sophisticated structures on prior, and 2) they do not properly capture label correlations with embeddings. To address the first, we learn a modality for each label class to form a mixture latent space. For the second, we replace the commonly used ranking and triplet losses with contrastive loss since contrastive loss involves more samples than triplet loss and has a larger capacity than ranking loss.

\begin{table*}[t]
\centering
\scalebox{0.775}{
\begin{tabular}{c@{\hspace{0.40em}}|c@{\hspace{0.40em}}c@{\hspace{0.40em}}c@{\hspace{0.40em}}c@{\hspace{0.40em}}c@{\hspace{0.40em}}c@{\hspace{0.40em}}c@{\hspace{0.40em}}c@{\hspace{0.40em}}c@{\hspace{0.40em}}||c@{\hspace{0.40em}}c@{\hspace{0.40em}}c@{\hspace{0.40em}}c@{\hspace{0.40em}}c@{\hspace{0.40em}}c@{\hspace{0.40em}}c@{\hspace{0.40em}}c@{\hspace{0.40em}}c@{\hspace{0.40em}}}
\toprule
Metric & \multicolumn{9}{c||}{example-F1} & \multicolumn{9}{c}{micro-F1} \\
\midrule
Dataset & \textit{eBird} & \textit{mirflickr} & \textit{nus-vec} & \textit{yeast} & \textit{scene} & \textit{sider} & \textit{reuters} & \textit{bkms} & \textit{delicious} &\textit{eBird} & \textit{mirflickr} & \textit{nus-vec} & \textit{yeast} & \textit{scene} & \textit{sider} & \textit{reuters} & \textit{bkms} & \textit{delicious}\\
\hline
BR & 0.365 & 0.325 & 0.343 & 0.630 & 0.606 & 0.766 & 0.733 & 0.171 & 0.174 & 0.384 & 0.371 & 0.371 & 0.655 & 0.706 & 0.796 & 0.767 & 0.125 & 0.197 \\
MLKNN & 0.510 & 0.383 & 0.342 & 0.618 & 0.691 & 0.738 & 0.703 & 0.213 & 0.259 & 0.557 & 0.415 & 0.368 & 0.625 & 0.667 & 0.772 & 0.680 & 0.181 & 0.264\\
HARAM & 0.510 & 0.432 & 0.396 & 0.629 & 0.717 & 0.722 & 0.711 & 0.216 & 0.267 & 0.573 & 0.447 & 0.415 & 0.635 & 0.693 & 0.754 & 0.695 & 0.230 & 0.273\\
SLEEC & 0.258 & 0.416 & 0.431 & 0.643 & 0.718 & 0.581 & 0.885 & 0.363 & 0.308 & 0.412 & 0.413 & 0.428 & 0.653 & 0.699 & 0.697 & 0.845 & 0.300 & 0.333\\
C2AE & 0.501 & 0.501 & 0.435 & 0.614 & 0.698 & 0.768 & 0.818 & 0.309 & 0.326 & 0.546 & 0.545 & 0.472 & 0.626 & 0.713 & 0.798 & 0.799 & 0.316 & 0.348\\
LaMP & 0.477 & 0.492 & 0.376 & 0.624 & 0.728 & 0.766 & \underline{0.906} & \underline{0.389} & 0.372 & 0.517 & 0.535 & 0.472 & 0.641 & 0.716 & 0.797 & \underline{0.886} & 0.373 & 0.386\\
MPVAE & \underline{0.551} & \underline{0.514} & 0.468 & \underline{0.648} & 0.751 & \underline{0.769} & 0.893 & 0.382 & \underline{0.373} & \underline{0.593} & \underline{0.552} & 0.492 & \underline{0.655} & 0.742 & \underline{0.800} & 0.881 & \underline{0.375} & \underline{0.393}\\
ASL & 0.528 & 0.477 & \underline{0.468} & 0.613 & \underline{0.770} & 0.752 & 0.880 & 0.373 & 0.359 & 0.580 & 0.525 & \underline{0.495} & 0.637 & \underline{0.753} & 0.795 & 0.869 & 0.354 & 0.387 \\
RBCC & 0.503 & 0.468 & 0.466 & 0.605 & 0.758 & 0.733 & 0.857 & - & - & 0.558 & 0.513 & 0.490 & 0.623 & 0.749 & 0.784 & 0.825 & - & - \\
\midrule
C-GMVAE & \textbf{0.576}  & \textbf{0.534}  & \textbf{0.481}  & \textbf{0.656}  & \textbf{0.777}  & \textbf{0.771}  & \textbf{0.917}  & \textbf{0.392} & \textbf{0.381} & \textbf{0.633}  & \textbf{0.575}  & \textbf{0.510}  & \textbf{0.665}  & \textbf{0.762}  & \textbf{0.803}  & \textbf{0.890}  & \textbf{0.377} & \textbf{0.403} \\
std ($\pm$) & 0.001 & 0.002 & 0.000 & 0.001 & 0.002 & 0.001 & 0.001 & 0.001 & 0.002 & 0.001 & 0.001 & 0.000 & 0.002 & 0.002 & 0.000 & 0.001 & 0.001 & 0.002\\
\bottomrule
\end{tabular}
}
\caption{The example-F1 (ex-F1) and micro-F1 (mi-F1) scores of different methods on all datasets. C-GMVAE's numbers are averaged over 3 seeds. The standard deviation (std) is also shown. 0.000 means an std$<0.0005$.}
\label{tab:ebf1_mif1}
\end{table*}

\begin{table*}[t]
\centering
\scalebox{0.775}{
\begin{tabular}{c@{\hspace{0.40em}}|c@{\hspace{0.40em}}c@{\hspace{0.40em}}c@{\hspace{0.40em}}c@{\hspace{0.40em}}c@{\hspace{0.40em}}c@{\hspace{0.40em}}c@{\hspace{0.40em}}c@{\hspace{0.40em}}c@{\hspace{0.40em}}||c@{\hspace{0.40em}}c@{\hspace{0.40em}}c@{\hspace{0.40em}}c@{\hspace{0.40em}}c@{\hspace{0.40em}}c@{\hspace{0.40em}}c@{\hspace{0.40em}}c@{\hspace{0.40em}}c@{\hspace{0.40em}}c@{\hspace{0.40em}}}
\toprule
Metric & \multicolumn{9}{c||}{macro-F1} & \multicolumn{9}{c}{Hamming Accuracy} \\
\midrule
Dataset & \textit{eBird} & \textit{mirflickr} & \textit{nus-vec} & \textit{yeast} & \textit{scene} & \textit{sider} & \textit{reuters} & \textit{bkms} & \textit{delicious} & \textit{eBird} & \textit{mirflickr} & \textit{nus-vec} & \textit{yeast} & \textit{scene} & \textit{sider} & \textit{reuters} & \textit{bkms} & \textit{delicious}\\
\hline
BR & 0.116 & 0.182 & 0.083 & 0.373 & 0.704 & 0.588 & 0.137 & 0.038 & 0.066 & 0.816 & 0.886 & 0.971 & 0.782 & 0.901 & 0.747 & 0.994 & 0.990 & 0.982 \\
MLKNN & 0.338 & 0.266 & 0.086 & 0.472 & 0.693 & 0.667 & 0.066 & 0.041 & 0.053 & 0.827 & 0.877 & 0.971 & 0.784 & 0.863 & 0.715 & 0.992 & 0.991 & 0.981 \\
HARAM & 0.474 & 0.284 & 0.157 & 0.448 & 0.713 & 0.649 & 0.100 & 0.140 & 0.074 & 0.819 & 0.634 & 0.971 & 0.744 & 0.902 & 0.650 & 0.905 & 0.990 & 0.981 \\
SLEEC & 0.363 & 0.364 & 0.135 & 0.425 & 0.699 & 0.592 & 0.403 & 0.195 & 0.142 & 0.816 & 0.870 & 0.971 & 0.782 & 0.894 & 0.675 & 0.996 & 0.989 & 0.982 \\
C2AE & 0.426 & 0.393 & 0.174 & 0.427 & 0.728 & 0.667 & 0.363 & 0.232 & 0.102 & 0.771 & 0.897 & 0.973 & 0.764 & 0.893 & 0.749 & 0.995 & 0.991 & 0.981 \\
LaMP & 0.381 & 0.387 & 0.203 & 0.480 & 0.745 & 0.668 & 0.520 & \underline{0.286} & \underline{0.196} & 0.811 & 0.897 & \underline{0.980} & 0.786 & 0.903 & 0.751 & 0.997 & \underline{0.992} & \underline{0.982} \\
MPVAE & \underline{0.494} & \underline{0.422} & \underline{0.211} & 0.482 & 0.750 & \underline{0.690} & 0.545 & 0.285 & 0.181 & 0.829 & \underline{0.898} & 0.980 & 0.792 & 0.909 & 0.755 & 0.997 & 0.991 & 0.982 \\
ASL & 0.467 & 0.410 & 0.208 & \underline{0.484} & \underline{0.765} & 0.668 & \underline{0.563} & 0.264 & 0.183 & \underline{0.831} & 0.893 & 0.975 & \underline{0.796} & \underline{0.912} & \underline{0.759} & \underline{0.997} & 0.991 & 0.982 \\
RBCC & 0.443 & 0.409 & 0.202 & 0.480 & 0.753 & 0.654 & 0.503 & - & - & 0.815 & 0.888 & 0.975 & 0.793 & 0.904 & 0.753 & 0.997 & - & - \\
\midrule
C-GMVAE & \textbf{0.538} & \textbf{0.440} & \textbf{0.226} & \textbf{0.487} & \textbf{0.769} & \textbf{0.691} & \textbf{0.582} & \textbf{0.291} & \textbf{0.197} & \textbf{0.847} & \textbf{0.903} & \textbf{0.984} & \textbf{0.796} & \textbf{0.915} & \textbf{0.767} & \textbf{0.997} & \textbf{0.992} & \textbf{0.983}\\
std ($\pm$) & 0.000 & 0.001 & 0.001 & 0.002 & 0.002 & 0.002 & 0.001 & 0.001 & 0.001 & 0.001 & 0.000 & 0.000 & 0.002 & 0.001 & 0.003 & 0.000 & 0.000 & 0.000\\
\bottomrule
\end{tabular}
}
\caption{The macro-F1 (ma-F1) and Hamming accuracy (HA) scores of different methods on all datasets. C-GMVAE's numbers are averaged over 3 seeds.}
\label{tab:maf1_ha}
\end{table*}

\begin{table*}[t]
\centering
\begin{tabular}{c|ccccccccc}
\toprule
Dataset & \textit{eBird} & \textit{mir.} & \textit{nus-vec} & \textit{yeast} & \textit{scene} & \textit{sider} & \textit{reuters} & \textit{bkms} & \textit{del.} \\
\midrule
BR & 0.598 & 0.582 & 0.443 & 0.745 & 0.700 & 0.573 & 0.752 & 0.301 & 0.485\\
MLKNN & 0.772 & 0.491 & 0.456 & 0.730 & 0.675 & 0.916 & 0.753 & 0.310 & 0.460\\
MLARAM & 0.768 & 0.350 & 0.404 & 0.682 & 0.722 & 0.930 & 0.679 & 0.312 & 0.419 \\
SLEEC & 0.656 & 0.623 & 0.531 & 0.745 & 0.730 & 0.882 & 0.908 & 0.415 & 0.676 \\
C2AE & 0.753 & 0.705 & 0.569 & 0.749 & 0.703 & 0.923 & 0.845 & 0.407 & 0.609\\
LaMP & 0.737 & 0.685 & 0.456 & 0.740 & 0.746 & 0.937 & 0.927 & 0.420 & 0.663\\
MPVAE & 0.820 & 0.726 & 0.587 & 0.743 & 0.777 & 0.958 & 0.930 & 0.437 & 0.696\\
ASL & 0.818 & 0.681 & 0.586 & 0.752 & 0.770 & 0.954 & 0.929 & 0.418 & 0.692 \\
RBCC & 0.805 & 0.682 & 0.582 & 0.745 & 0.777 & 0.942 & 0.913 & - & - \\
\midrule
C-GMVAE & \textbf{0.825} & \textbf{0.732} & \textbf{0.595} & \textbf{0.751} & \textbf{0.788} & \textbf{0.962} & \textbf{0.939} & \textbf{0.465} & \textbf{0.707} \\
\bottomrule
\end{tabular}
\caption{The precision@1 scores of different methods on all datasets. ``mir." stands for mirflickr and ``del." means delicious dataset.}
\label{tab:p_at_1}
\end{table*}

\section{Related Work}
Learning a shared latent space for features and labels is a common and useful idea. For single-label prediction tasks, CADA-VAE \cite{schonfeld2019generalized} learns and aligns latent label and feature spaces through distribution alignment losses. Similar ideas can be seen in out-of-distribution detection as well \cite{sundar2020out}. In multi-label scenarios, methods adopting this idea typically have a similar module that directly maps the multi-hot labels to embeddings \cite{yeh2017learning,chen2019two,bai2020disentangled}. This is a rather difficult learning task. Suppose we have 30 label categories. There could be up to $2^{30}$ label sets. For probabilistic models like MPVAE, 
that means one latent label space has to represent up to $2^{30}$ label combinations.
In contrast, C-GMVAE  learns per-category subspaces and 
forms a mixture prior distribution based on the observed samples' label sets.

Contrastive learning has become one of the most popular self-supervised learning techniques. It has also been applied to supervised learning tasks. SupCon \cite{khosla2020supervised} first demonstrated the effectiveness of supervised contrastive loss in image classification tasks. It was soon generalized to other domains like visual reasoning \cite{malkinski2020multi, dao2021contrast}. 
Nevertheless, these methods depend on vision-specific augmentation techniques and attention mechanisms. 
Another related work is multi-label contrastive learning \cite{song2020multi}. 
But the work does not deal with MLC. Instead, it extends contrastive learning to the identification of more than one positive sample, which resembles a multi-label scenario.

Some earlier works also attempted metric learning or triplet loss in MLC \cite{annarumma2017deep}. Triplet loss typically only takes one pair of positive and negative samples for one anchor, while contrastive loss uses many more negative and positive samples. Recent papers found that more samples can greatly boost performance \cite{chen2020simple,wang2020unsupervised}. Though our contrastive module is constrained by the maximum number of label classes, the number of used samples has already surpassed other losses (e.g., triplet loss), and our observations reinforce that more samples help with the performance.

\section{Experiments}

\begin{table*}[t]
\centering
\begin{tabular}{c|cccccc}
\toprule
Dataset &
  \# Samples &
  \# Labels &
  \begin{tabular}[c]{@{}c@{}}Mean\\ Labels\\ /Sample\end{tabular} &
  \begin{tabular}[c]{@{}c@{}}Median\\ Labels\\ /Sample\end{tabular} &
  \begin{tabular}[c]{@{}c@{}}Max\\ Labels\\ /Sample\end{tabular} &
  \begin{tabular}[c]{@{}c@{}}Mean\\ Samples\\ /Label\end{tabular} \\
\midrule
\textit{eBird}     & 41778  & 100 & 20.69 & 18 & 96 & 8322.95 \\
\textit{bookmarks} & 87856  & 208 & 2.03  & 1  & 44 & 584.67  \\
\textit{nus-vec}   & 269648 & 85  & 1.86  & 1  & 12 & 3721.7  \\
\textit{mirflickr} & 25000  & 38  & 4.80  & 5  & 17 & 1247.34   \\
\textit{reuters}   & 10789   & 90  & 1.23  & 1  & 15 & 106.50    \\
\textit{scene}     & 2407   & 6   & 1.07  & 1  & 3  & 170.83    \\
\textit{sider}     & 1427   & 27  & 15.3  & 16 & 26 & 731.07  \\
\textit{yeast}     & 2417   & 14  & 4.24  & 4  & 11 & 363.14  \\
\textit{delicious} & 16105  & 983 & 19.06 & 20 & 25 & 250.15 \\
\bottomrule
\end{tabular}
\caption{Dataset Statistics.}
\label{tab:dataset_stats}
\end{table*}

\begin{table}[t]
\centering
\begin{tabular}{c|c|ccc}
\toprule
& variations & eb-F1 & mi-F1 & ma-F1 \\
\midrule
\multirow{3}{*}{\textit{ebird}} & uni-Gaussian  & 0.545 & 0.583 & 0.490 \\
 & GM only  & 0.561 & 0.603 & 0.511 \\
 & contrastive only & 0.558 & 0.594 & 0.515 \\
 & GM+contrastive & 0.576 & 0.633 & 0.538 \\
\hline
\multirow{3}{*}{\textit{mirflickr}} & uni-Gaussian &  0.510 & 0.541 & 0.413 \\
& GM only &  0.521 & 0.561 & 0.429 \\
 & contrastive only & 0.526 & 0.565 & 0.428 \\
 & GM+contrastive  & 0.534 & 0.575 & 0.440 \\
\hline
\multirow{3}{*}{\textit{nus-vec}} & uni-Gaussian & 0.461 & 0.479 & 0.203 \\
& GM only & 0.472 & 0.505 & 0.218 \\
 & contrastive only & 0.470 & 0.501 & 0.213 \\
 & GM+contrastive  & 0.481 & 0.510 & 0.226 \\
\bottomrule
\end{tabular}
\caption{Ablation study on the contrastive learning module and the Gaussian mixture module. Note that both modules are contributions of this work. As shown in the table, GM consistently improves performance. The contrastive module can also further boost the performance.}
\label{tab:ablation}
\end{table}

\begin{table}[t]
\centering
\scalebox{0.93}{
\begin{tabular}{c@{\hspace{0.45em}}|c@{\hspace{0.45em}}|c@{\hspace{0.45em}}c@{\hspace{0.45em}}c@{\hspace{0.45em}}c@{\hspace{0.45em}}}
\toprule
 & method (data \%) & HA & ex-F1 & mi-F1 & ma-F1 \\
\midrule
\multirow{2}{*}{\textit{ebird}} & MPVAE (100\%) & 0.829 & 0.551 & 0.593 & 0.494 \\
 & C-GMVAE (50\%) & 0.842 & 0.557 & 0.615 & 0.521 \\ \hline
\multirow{2}{*}{\textit{mirflickr}} & MPVAE (100\%) & 0.898 & 0.514 & 0.552 & 0.422 \\
 & C-GMVAE (50\%) & 0.899 & 0.512 & 0.553 & 0.412 \\ \hline
\multirow{2}{*}{\textit{nus-vec}} & MPVAE (100\%) & 0.980 & 0.468 & 0.492 & 0.211 \\
 & C-GMVAE (50\%) & 0.975 & 0.465 & 0.494 & 0.201\\
\bottomrule
\end{tabular}
}
\caption{Comparisons between MPVAE and C-GMVAE using 100\% and 50\% respectively.}
\label{tab:semi}
\end{table}

\begin{table*}[t]
\centering
\begin{tabular}{c|cccccc}
\toprule
module in C-GMVAE & \textit{eBird} & \textit{mirflickr} & \textit{nus-vec} & \textit{yeast} & \textit{scene} & \textit{sider} \\
\midrule
Covariance & 0.601  & 0.556  & 0.488  & 0.650 & 0.751 & 0.787 \\
GNN & 0.599 & 0.560 & 0.491 & 0.655 & 0.749 & 0.801\\
\midrule
contrastive & \textbf{0.633}  & \textbf{0.575}  & \textbf{0.510}  & \textbf{0.665} & \textbf{0.762} & \textbf{0.803} \\
\bottomrule
\end{tabular}
\caption{mi-F1 performance after replacing our contrastive module with a GNN or a covariance matrix.}
\label{tab:abl_replace}
\end{table*}

We have various setups to validate the performance of C-GMVAE. First, we compare the example-F1, micro-F1 and macro-F1 scores, Hamming accuracies and precision@1 of different methods. Second, we compare their performance when fewer training data are available. Third, an ablation study shows the importance of the proposed modules. Finally, we demonstrate the interpretability of the label embeddings on the \textit{eBird} dataset. Our code is publicly available\footnote{https://github.com/JunwenBai/c-gmvae}.

\subsection{Setup}
For the main evaluation experiments, we use nine datasets, including image datasets \textit{mirflickr}, \textit{nuswide}, \textit{scene}~\cite{huiskes2008mir,chua2009nus,boutell2004learning}, biology datasets \textit{sider}, \textit{yeast}~\cite{kuhn2016sider,nakai1992knowledge}, ecology dataset \textit{eBird}~\cite{fink2017ebird},  and text datasets \textit{reuters, bookmarks, delicious}~\cite{debole2005analysis,katakis2008multilabel,tsoumakas2008effective} (see Tab.~\ref{tab:dataset_stats} for dataset statistics). All features are collected in vector format \cite{lanchantin2019neural,bai2020disentangled}.
The feature pre-processing is standard following previous works \cite{lanchantin2019neural,bai2020disentangled} and the datasets are public\footnote{http://mulan.sourceforge.net/datasets-mlc.html}. 
Each dataset is separated into training (80\%), validation (10\%) and testing (10\%) splits. The datasets are also preprocessed to fit the input formats of different methods. We use mini-batch training with batch size 128. Each batch is randomly sampled from the dataset.

The evaluation metrics are three F1 scores, Hamming accuracy and precision@1. The evaluation process, model selection and preprocessing strictly follow previous works \cite{tu2018learning,lanchantin2019neural,bai2020disentangled}. Most numbers are also directly quoted from the corresponding papers for comparison. 
Our method is compared against ASL \cite{ridnik2021asymmetric}, RBCC \cite{gerych2021recurrent}, MPVAE \cite{bai2020disentangled}, LaMP \cite{lanchantin2019neural}, C2AE \cite{yeh2017learning}, SLEEC \cite{bhatia2015sparse}, HARAM \cite{benites2015haram}, MLKNN \cite{zhang2007ml}, and BR \cite{zhang2018binary}. 

ASL introduces asymmetric loss, a variant of BCE and focal loss for MLC, and requires tuning of focusing parameters. RBCC is based on a Bayesian network, which requires structure learning to derive a directed acyclic graph (DAG) first. MPVAE is a novel method which learns and aligns the probabilistic feature and label subspaces. Label correlations are captured by a multivariate probit module. LaMP  adopts attention-based neural message passing to handle the label correlations, which is a neural extension of previous CRF-based methods. C2AE was one of the first papers to use NNs to learn and align latent spaces. C2AE imposes a canonical correlation analysis (CCA) constraint on the latent space. SLEEC explores the low-rank assumption in MLC, to reduce the effective number of labels. Other deep methods make similar  low-rank assumptions. HARAM  was one of the first methods to introduced NNs to MLC. MLKNN is a classic MLC method using k-nearest neighbors (KNN). It finds nearest examples to a test sample and adopts Bayesian inference to select assigned labels. Lastly, binary relevance (BR) is one of the most intuitive solutions for MLC, which decomposes the multi-label scenario into independent binary prediction tasks.

\subsection{Metrics}
We evaluate our method trained with objective Eq.~\ref{eq:final_obj} on several commonly used multi-label metrics. Suppose the ground-truth label is $y$ and the predicted label is $\hat{y}$. We denote true positives, false positives, false negatives by $tp_j, fp_j, fn_j$ respectively for the $j$-th of $L$ label categories. 
(i) HA: $\frac{1}{L}\sum_{j=1}^L\mathbbm{1}[y_j=\hat{y}_j]$
(ii) example-F1: $\frac{2\sum_{j=1}^Ly_i\hat{y}_i}{\sum_{j=1}^Ly_i + \sum_{j=1}^L\hat{y}_i}$
(iii) micro-F1:
    $\frac{\sum_{j=1}^Ltp_j}{\sum_{j=1}^L 2tp_j+fp_j+fn_j}$
(iv) macro-F1:
    $\frac{1}{L}\sum_{j=1}^L\frac{2tp_j}{2tp_j+fp_j+fn_j}$

Furthermore, precision@1 is the proportion of correctly predicted labels in the top-1 predictions.

\subsection{Architecture and Hyperparameters}
As we state in the introduction, we do not require very sophisticated neural architectures in C-GMVAE. All the neural layers are fully connected. 
The feature encoder is a fully connected NN with 3 hidden layers and the activation function is ReLU. The label encoder is also fully connected comprising two hidden layers and the decoder has two hidden layers as well. More details of the model can be found in the appendix.
We set $\alpha=1,\beta=0.5,E=2048$ by default. Grid search is applied to find the best learning rate, dropout ratio, and weight decay ratio for each dataset.
We use one Nvidia V100 GPU for all experiments.
More architecture, hyper-parameter tuning, and implementation details can be found in the appendix.

\subsection{Evaluations}

\begin{figure}[t]
\centering
\includegraphics[width=\linewidth]{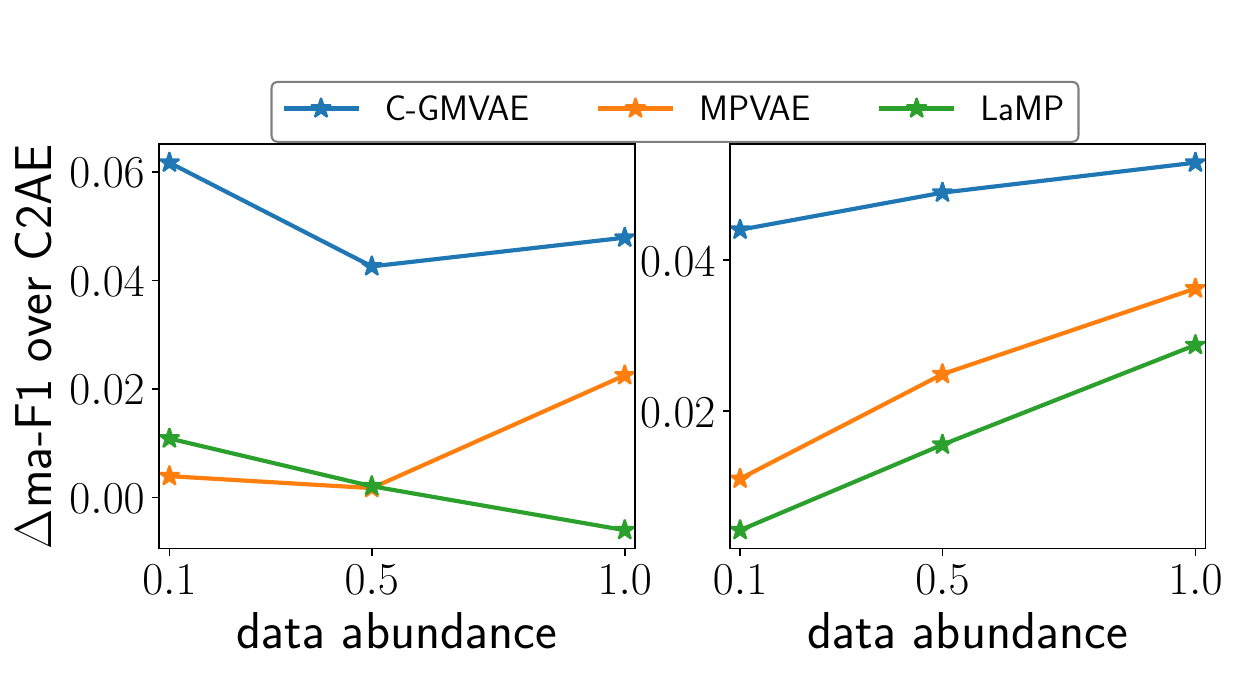}
\caption{Relative improvements of C-GMVAE, MPVAE and LaMP on ma-F1 compared to C2AE. Left and right plots correspond to \textit{mirflickr} and \textit{nus-vec} datasets respectively. Every method (including C2AE) is trained on the same amount of data (10\%, 50\% or 100\%) for comparison.}
\label{fig:semi}
\end{figure}

\paragraph{Full supervision} 
In the full supervision scenario, which is commonly adopted by the methods we compare against, we evaluate five metrics: example-F1 (ex-F1), micro-F1 (mi-F1), macro-F1 (ma-F1), Hamming accuracy (HA) and precision@1. The ex-F1 score is the averaged F1-score over all the samples. The mi-F1 score measures the aggregated contributions of all classes. The ma-F1 treats each class equally and takes the class-wise average. HA counts the correctly predicted labels regardless of samples or classes. The full definitions of these metrics can be found in the appendix.

Tab.~\ref{tab:ebf1_mif1}, \ref{tab:maf1_ha} and \ref{tab:p_at_1} present the performance of all the methods w.r.t. the metrics. We abbreviate \textit{nuswide-vector} to \textit{nus-vec}, and \textit{bookmarks} to \textit{bkms}. C-GMVAE outperforms the existing state-of-the-art methods on all the datasets. The best numbers are marked in bold. All the numbers for C-GMVAE are averaged over 3 seeds for stability and the standard deviations are included in the table. On ex-F1, C-GMVAE improves over ASL by 5.3\%, RBCC by 7.7\%, MPVAE by 2.5\%, and LaMP by 8.8\% on average across all the datasets. Similarly, on mi-F1, C-GMVAE improves over ASL by 4.4\%, RBCC by 6.7\%, MPVAE by 2.4\% and LaMP 6.1\% on average. On ma-F1, the improvements are as large as 6.1\%, 9.4\%, 4.1\% and 11\%, respectively. 
C-GMVAE outperforms other methods consistently.

\paragraph{Ablation study} To demonstrate the strength of C-GMVAE, we compare it with a unimodal Gaussian latent model, a Gaussian mixture only latent model (without contrastive module), and a contrastive learning only model (without the KL divergence term) in Tab.~\ref{tab:ablation}. Our C-GMVAE (GM+contrastive) consistently outperforms other models by a large margin. For instance, on ma-F1, C-GMVAE improves over the unimodal Gaussian model by 7\%. In Tab.~\ref{tab:abl_replace}, we show that the contrastive module outperforms both the GNN and covariance matrix modules.


\begin{figure}[t]
\centering
\includegraphics[width=\linewidth]{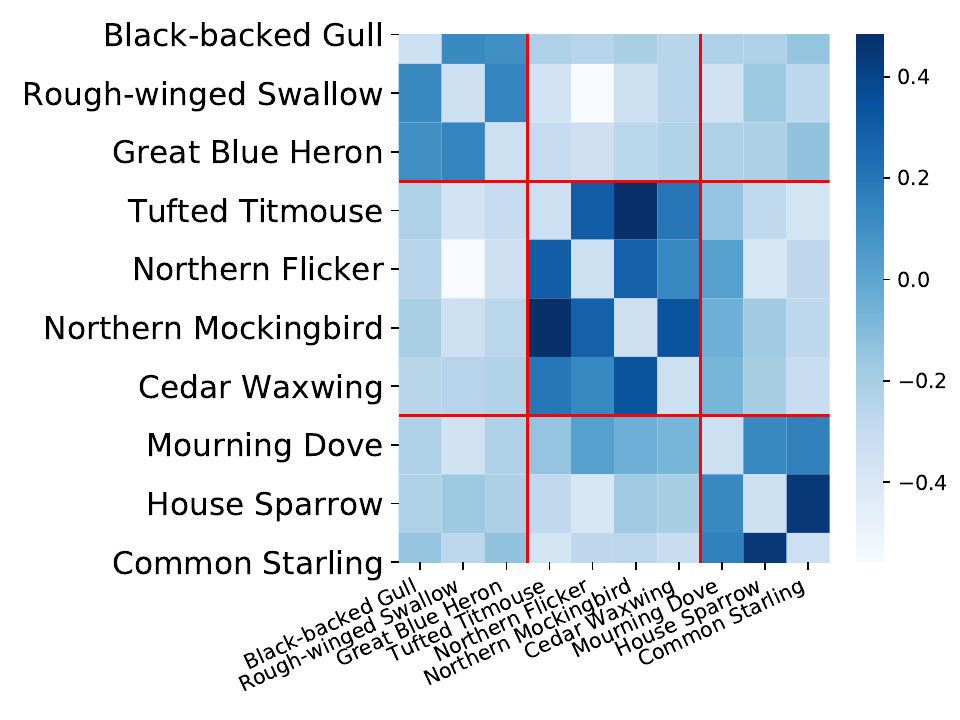}
\caption{Label-label inner-products from C-GMVAE. One can compare it with Fig.~\ref{fig:bird_mpvae} from MPVAE in the appendix. C-GMVAE demonstrates sharper and more meaningful inner-products.}
\label{fig:bird}
\end{figure}

\paragraph{Training on fewer data} Contrastive learning learns contrastive views and thus requires less information compared to generative learning, which demands a more complete representation for reconstruction. Contrastive learning has the potential to discover the intrinsic structure present in the data, and therefore is widely used in self-supervised learning because it generalizes well. We observe this with C-GMVAE as well. To demonstrate this, we shrink the size of training data by 50\% or 90\% and train methods on them. Surprisingly, we find C-GMVAE can often match the performance of other methods with only 50\% of the training data.
Tab.~\ref{tab:semi} compares MPVAE trained on all data and C-GMVAE trained on 50\% of the data. Their performance is approximately the same. We further compare several major state-of-the-art methods including ours, all trained on the same \textbf{randomly} selected 10\%, 50\% and 100\% of the data, and show their performance over C2AE. Fig.~\ref{fig:semi} shows the improvements over C2AE on ma-F1. Ours clearly outperforms the others with fewer data. More plots for other datasets and metrics are in the appendix.

\paragraph{Interpretability} Our work is also motivated by ecological applications~\cite{gomes2019computational}, where it is important to understand species interactions. Fig.~\ref{fig:bird} shows a map of inner-product weights of label embeddings on the eBird dataset. The bird species on the x-axis and the y-axis are the same. The first 3 bird species are water birds. The following 4 bird species are forest birds. The last 3 bird species are residential birds. 
Darker colors indicate more similar birds. We subtract the diagonal to exclude the self correlation. 
The heatmap matrix clearly forms three blocks on the diagonal. The first block contains Black-backed Gull, Rough-winged Swallow and Great Blue Heron. These three birds are water birds living near sea or lake. The second block has Tufted Titmouse, Northern Flicker, Northern Mockingbird, and Cedar Waxwing. These birds typically live in the forest with a lot of trees. The remaining birds are commonly seen residential birds, Mourning Dove, House Sparrow and Common Starling. They live inside or near human residences. Since human activities are wide-spread, the distribution of these birds is therefore quite broad. For example, the Mourning Dove is also correlated with forest birds in Fig.~\ref{fig:bird}. But one can observe that for each group of birds, their intra-group correlations are always stronger than inter-group correlations.
Therefore, the learnt embeddings do encompass semantic meanings. The derived correlations could also help the study of wildlife protection \cite{johnston2019best}. 

\section{Conclusion}

In this work, we propose the contrastive learning boosted Gaussian mixture variational autoencoder (C-GMVAE), 
a novel method for multi-label prediction tasks. C-GMVAE combines 
the learning of Gaussian mixture latent spaces and the contrastive learning of feature and label embeddings. Not only does C-GMVAE achieve the state-of-the-art performance, it also provides insights into semi-supervised learning and model interpretability. Interesting future directions include the exploration of various contrastive learning mechanisms, model architecture improvements, and other latent space structures. 

\section*{Acknowledgement}

Our work is supported by NSF Expedition CompSustNet CCF-1522054, NSF Computer and Network Systems grant CNS-1059284, and Defense University Research Instrumentation Program ARO DURIP W911NF-17-1-0187. We thank Rich Bernstein for proofreading.


\bibliography{ref}

\begin{thebibliography}{58}
\providecommand{\natexlab}[1]{#1}
\providecommand{\url}[1]{\texttt{#1}}
\expandafter\ifx\csname urlstyle\endcsname\relax
  \providecommand{\doi}[1]{doi: #1}\else
  \providecommand{\doi}{doi: \begingroup \urlstyle{rm}\Url}\fi

\bibitem[Annarumma \& Montana(2017)Annarumma and Montana]{annarumma2017deep}
Annarumma, M. and Montana, G.
\newblock Deep metric learning for multi-labelled radiographs.
\newblock \emph{arXiv preprint arXiv:1712.07682}, 2017.

\bibitem[Bai et~al.(2020)Bai, Kong, and Gomes]{bai2020disentangled}
Bai, J., Kong, S., and Gomes, C.
\newblock Disentangled variational autoencoder based multi-label classification
  with covariance-aware multivariate probit model.
\newblock \emph{IJCAI}, 2020.

\bibitem[Belanger \& McCallum(2016)Belanger and
  McCallum]{belanger2016structured}
Belanger, D. and McCallum, A.
\newblock Structured prediction energy networks.
\newblock In \emph{International Conference on Machine Learning}, 2016.

\bibitem[Benites \& Sapozhnikova(2015)Benites and
  Sapozhnikova]{benites2015haram}
Benites, F. and Sapozhnikova, E.
\newblock Haram: a hierarchical aram neural network for large-scale text
  classification.
\newblock In \emph{2015 IEEE international conference on data mining workshop
  (ICDMW)}. IEEE, 2015.

\bibitem[Bhatia et~al.(2015)Bhatia, Jain, Kar, Varma, and
  Jain]{bhatia2015sparse}
Bhatia, K., Jain, H., Kar, P., Varma, M., and Jain, P.
\newblock Sparse local embeddings for extreme multi-label classification.
\newblock \emph{Advances in neural information processing systems},
  28:\penalty0 730--738, 2015.

\bibitem[Bi \& Kwok(2014)Bi and Kwok]{bi2014multilabel}
Bi, W. and Kwok, J.
\newblock Multilabel classification with label correlations and missing labels.
\newblock In \emph{Proceedings of the AAAI Conference on Artificial
  Intelligence}, 2014.

\bibitem[Boutell et~al.(2004)Boutell, Luo, Shen, and
  Brown]{boutell2004learning}
Boutell, M.~R., Luo, J., Shen, X., and Brown, C.~M.
\newblock Learning multi-label scene classification.
\newblock \emph{Pattern recognition}, 37\penalty0 (9):\penalty0 1757--1771,
  2004.

\bibitem[Chang et~al.(2019)Chang, Yu, Zhong, Yang, and Dhillon]{chang2019x}
Chang, W.-C., Yu, H.-F., Zhong, K., Yang, Y., and Dhillon, I.
\newblock X-bert: extreme multi-label text classification with using
  bidirectional encoder representations from transformers.
\newblock \emph{arXiv preprint arXiv:1905.02331}, 2019.

\bibitem[Chen et~al.(2019{\natexlab{a}})Chen, Wang, Liu, Zhao, Hu, and
  Chen]{chen2019two}
Chen, C., Wang, H., Liu, W., Zhao, X., Hu, T., and Chen, G.
\newblock Two-stage label embedding via neural factorization machine for
  multi-label classification.
\newblock In \emph{AAAI}, 2019{\natexlab{a}}.

\bibitem[Chen et~al.(2020)Chen, Kornblith, Norouzi, and Hinton]{chen2020simple}
Chen, T., Kornblith, S., Norouzi, M., and Hinton, G.
\newblock A simple framework for contrastive learning of visual
  representations.
\newblock \emph{arXiv preprint arXiv:2002.05709}, 2020.

\bibitem[Chen et~al.(2019{\natexlab{b}})Chen, Wei, Wang, and
  Guo]{chen2019multi}
Chen, Z.-M., Wei, X.-S., Wang, P., and Guo, Y.
\newblock Multi-label image recognition with graph convolutional networks.
\newblock In \emph{Proceedings of the IEEE/CVF Conference on Computer Vision
  and Pattern Recognition}, pp.\  5177--5186, 2019{\natexlab{b}}.

\bibitem[Chua et~al.(2009)Chua, Tang, Hong, Li, Luo, and Zheng]{chua2009nus}
Chua, T.-S., Tang, J., Hong, R., Li, H., Luo, Z., and Zheng, Y.
\newblock Nus-wide: a real-world web image database from national university of
  singapore.
\newblock In \emph{Proceedings of the ACM international conference on image and
  video retrieval}, pp.\  1--9, 2009.

\bibitem[Chung et~al.(2015)Chung, Kastner, Dinh, Goel, Courville, and
  Bengio]{chung2015recurrent}
Chung, J., Kastner, K., Dinh, L., Goel, K., Courville, A.~C., and Bengio, Y.
\newblock A recurrent latent variable model for sequential data.
\newblock \emph{Advances in neural information processing systems},
  28:\penalty0 2980--2988, 2015.

\bibitem[Dao et~al.(2021)Dao, Ethan, Dinh, and Jianfei]{dao2021contrast}
Dao, S.~D., Ethan, Z., Dinh, P., and Jianfei, C.
\newblock Contrast learning visual attention for multi label classification.
\newblock \emph{arXiv preprint arXiv:2107.11626}, 2021.

\bibitem[Debole \& Sebastiani(2005)Debole and Sebastiani]{debole2005analysis}
Debole, F. and Sebastiani, F.
\newblock An analysis of the relative hardness of reuters-21578 subsets.
\newblock \emph{Journal of the American Society for Information Science and
  technology}, 56\penalty0 (6):\penalty0 584--596, 2005.

\bibitem[Decubber et~al.(2018)Decubber, Mortier, Dembczy{\'n}ski, and
  Waegeman]{decubber2018deep}
Decubber, S., Mortier, T., Dembczy{\'n}ski, K., and Waegeman, W.
\newblock Deep f-measure maximization in multi-label classification: A
  comparative study.
\newblock In \emph{Joint European Conference on Machine Learning and Knowledge
  Discovery in Databases}, pp.\  290--305. Springer, 2018.

\bibitem[Dilokthanakul et~al.(2016)Dilokthanakul, Mediano, Garnelo, Lee,
  Salimbeni, Arulkumaran, and Shanahan]{dilokthanakul2016deep}
Dilokthanakul, N., Mediano, P.~A., Garnelo, M., Lee, M.~C., Salimbeni, H.,
  Arulkumaran, K., and Shanahan, M.
\newblock Deep unsupervised clustering with gaussian mixture variational
  autoencoders.
\newblock \emph{arXiv preprint arXiv:1611.02648}, 2016.

\bibitem[Doersch(2016)]{doersch2016tutorial}
Doersch, C.
\newblock Tutorial on variational autoencoders.
\newblock \emph{arXiv preprint arXiv:1606.05908}, 2016.

\bibitem[Eslami et~al.(2016)Eslami, Heess, Weber, Tassa, Szepesvari,
  Kavukcuoglu, and Hinton]{eslami2016attend}
Eslami, S., Heess, N., Weber, T., Tassa, Y., Szepesvari, D., Kavukcuoglu, K.,
  and Hinton, G.~E.
\newblock Attend, infer, repeat: Fast scene understanding with generative
  models.
\newblock \emph{arXiv preprint arXiv:1603.08575}, 2016.

\bibitem[Fink et~al.(2017)Fink, Auer, Obregon, Hochachka, Iliff, Sullivan,
  Wood, Davies, and Kelling]{fink2017ebird}
Fink, D., Auer, T., Obregon, F., Hochachka, W., Iliff, M., Sullivan, B., Wood,
  C., Davies, I., and Kelling, S.
\newblock The ebird reference dataset version 2016 (erd2016), 2017.

\bibitem[Gerych et~al.(2021)Gerych, Hartvigsen, Buquicchio, Agu, and
  Rundensteiner]{gerych2021recurrent}
Gerych, W., Hartvigsen, T., Buquicchio, L., Agu, E., and Rundensteiner, E.~A.
\newblock Recurrent bayesian classifier chains for exact multi-label
  classification.
\newblock \emph{Advances in Neural Information Processing Systems},
  34:\penalty0 15981--15992, 2021.

\bibitem[Gomes et~al.(2019)Gomes, Dietterich, et~al.]{gomes2019computational}
Gomes, C., Dietterich, T., et~al.
\newblock Computational sustainability: Computing for a better world and a
  sustainable future.
\newblock \emph{Communications of the ACM}, 62\penalty0 (9):\penalty0 56--65,
  2019.

\bibitem[Gutmann \& Hyv{\"a}rinen(2010)Gutmann and
  Hyv{\"a}rinen]{gutmann2010noise}
Gutmann, M. and Hyv{\"a}rinen, A.
\newblock Noise-contrastive estimation: A new estimation principle for
  unnormalized statistical models.
\newblock In \emph{AISTATS}, 2010.

\bibitem[Huiskes \& Lew(2008)Huiskes and Lew]{huiskes2008mir}
Huiskes, M.~J. and Lew, M.~S.
\newblock The mir flickr retrieval evaluation.
\newblock In \emph{Proceedings of the 1st ACM international conference on
  Multimedia information retrieval}, pp.\  39--43, 2008.

\bibitem[Johnston et~al.(2019)Johnston, Hochachka, Strimas-Mackey, Gutierrez,
  Robinson, Miller, Auer, Kelling, and Fink]{johnston2019best}
Johnston, A., Hochachka, W., Strimas-Mackey, M., Gutierrez, V.~R., Robinson,
  O., Miller, E., Auer, T., Kelling, S., and Fink, D.
\newblock Best practices for making reliable inferences from citizen science
  data: case study using ebird to estimate species distributions.
\newblock \emph{BioRxiv}, pp.\  574392, 2019.

\bibitem[Katakis et~al.(2008)Katakis, Tsoumakas, and
  Vlahavas]{katakis2008multilabel}
Katakis, I., Tsoumakas, G., and Vlahavas, I.
\newblock Multilabel text classification for automated tag suggestion.
\newblock \emph{ECML PKDD Discovery Challenge}, pp.\ ~75, 2008.

\bibitem[Khosla et~al.(2021)Khosla, Teterwak, Wang, Sarna, Tian, Isola,
  Maschinot, Liu, and Krishnan]{khosla2020supervised}
Khosla, P., Teterwak, P., Wang, C., Sarna, A., Tian, Y., Isola, P., Maschinot,
  A., Liu, C., and Krishnan, D.
\newblock Supervised contrastive learning.
\newblock \emph{arXiv preprint arXiv:2004.11362}, 2021.

\bibitem[Kingma \& Ba(2014)Kingma and Ba]{kingma2014adam}
Kingma, D.~P. and Ba, J.
\newblock Adam: A method for stochastic optimization.
\newblock \emph{arXiv:1412.6980}, 2014.

\bibitem[Kingma \& Welling(2013)Kingma and Welling]{kingma2013auto}
Kingma, D.~P. and Welling, M.
\newblock Auto-encoding variational bayes.
\newblock \emph{arXiv:1312.6114}, 2013.

\bibitem[Koyejo et~al.(2015)Koyejo, Natarajan, Ravikumar, and
  Dhillon]{koyejo2015consistent}
Koyejo, O., Natarajan, N., Ravikumar, P., and Dhillon, I.~S.
\newblock Consistent multilabel classification.
\newblock In \emph{NIPS}, volume~29, pp.\  3321--3329, 2015.

\bibitem[Kuhn et~al.(2016)Kuhn, Letunic, Jensen, and Bork]{kuhn2016sider}
Kuhn, M., Letunic, I., Jensen, L.~J., and Bork, P.
\newblock The sider database of drugs and side effects.
\newblock \emph{Nucleic acids research}, 44\penalty0 (D1):\penalty0
  D1075--D1079, 2016.

\bibitem[Lanchantin et~al.(2019)Lanchantin, Sekhon, and
  Qi]{lanchantin2019neural}
Lanchantin, J., Sekhon, A., and Qi, Y.
\newblock Neural message passing for multi-label classification.
\newblock \emph{arXiv preprint arXiv:1904.08049}, 2019.

\bibitem[Ma{\l}ki{\'n}ski \& Ma{\'n}dziuk(2020)Ma{\l}ki{\'n}ski and
  Ma{\'n}dziuk]{malkinski2020multi}
Ma{\l}ki{\'n}ski, M. and Ma{\'n}dziuk, J.
\newblock Multi-label contrastive learning for abstract visual reasoning.
\newblock \emph{arXiv preprint arXiv:2012.01944}, 2020.

\bibitem[Mikolov et~al.(2013)Mikolov, Chen, Corrado, and
  Dean]{mikolov2013efficient}
Mikolov, T., Chen, K., Corrado, G., and Dean, J.
\newblock Efficient estimation of word representations in vector space.
\newblock \emph{arXiv preprint arXiv:1301.3781}, 2013.

\bibitem[Nakai \& Kanehisa(1992)Nakai and Kanehisa]{nakai1992knowledge}
Nakai, K. and Kanehisa, M.
\newblock A knowledge base for predicting protein localization sites in
  eukaryotic cells.
\newblock \emph{Genomics}, 14\penalty0 (4):\penalty0 897--911, 1992.

\bibitem[Oord et~al.(2018)Oord, Li, and Vinyals]{oord2018representation}
Oord, A. v.~d., Li, Y., and Vinyals, O.
\newblock Representation learning with contrastive predictive coding.
\newblock \emph{arXiv preprint arXiv:1807.03748}, 2018.

\bibitem[Read et~al.(2009)Read, Pfahringer, Holmes, and
  Frank]{read2009classifier}
Read, J., Pfahringer, B., Holmes, G., and Frank, E.
\newblock Classifier chains for multi-label classification.
\newblock In \emph{Joint European Conference on Machine Learning and Knowledge
  Discovery in Databases}. Springer, 2009.

\bibitem[Ridnik et~al.(2021)Ridnik, Ben-Baruch, Zamir, Noy, Friedman, Protter,
  and Zelnik-Manor]{ridnik2021asymmetric}
Ridnik, T., Ben-Baruch, E., Zamir, N., Noy, A., Friedman, I., Protter, M., and
  Zelnik-Manor, L.
\newblock Asymmetric loss for multi-label classification.
\newblock In \emph{Proceedings of the IEEE/CVF International Conference on
  Computer Vision}, pp.\  82--91, 2021.

\bibitem[Sch{\"o}nfeld et~al.(2019)Sch{\"o}nfeld, Ebrahimi, Sinha, Darrell, and
  Akata]{schonfeld2019generalized}
Sch{\"o}nfeld, E., Ebrahimi, S., Sinha, S., Darrell, T., and Akata, Z.
\newblock Generalized zero-and few-shot learning via aligned variational
  autoencoders.
\newblock In \emph{CVPR}. IEEE, 2019.

\bibitem[Seymour \& Zhang(2018)Seymour and Zhang]{seymour2018multi}
Seymour, Z. and Zhang, Z.
\newblock Multi-label triplet embeddings for image annotation from
  user-generated tags.
\newblock In \emph{Proceedings of the 2018 ACM on International Conference on
  Multimedia Retrieval}, pp.\  249--256, 2018.

\bibitem[Shi et~al.(2019)Shi, Siddharth, Paige, and Torr]{shi2019variational}
Shi, Y., Siddharth, N., Paige, B., and Torr, P.~H.
\newblock Variational mixture-of-experts autoencoders for multi-modal deep
  generative models.
\newblock \emph{arXiv preprint arXiv:1911.03393}, 2019.

\bibitem[Shi et~al.(2020)Shi, Paige, Torr, and Siddharth]{shi2020relating}
Shi, Y., Paige, B., Torr, P.~H., and Siddharth, N.
\newblock Relating by contrasting: A data-efficient framework for multimodal
  generative models.
\newblock \emph{arXiv preprint arXiv:2007.01179}, 2020.

\bibitem[Shu(2016)]{shu2016gaussian}
Shu, R.
\newblock Gaussian mixture vae: Lessons in variational inference, generative
  models, and deep nets.
\newblock 2016.

\bibitem[Song \& Ermon(2020)Song and Ermon]{song2020multi}
Song, J. and Ermon, S.
\newblock Multi-label contrastive predictive coding.
\newblock \emph{Advances in Neural Information Processing Systems}, 33, 2020.

\bibitem[Sundar et~al.(2020)Sundar, Ramakrishna, Rahiminasab, Easwaran, and
  Dubey]{sundar2020out}
Sundar, V.~K., Ramakrishna, S., Rahiminasab, Z., Easwaran, A., and Dubey, A.
\newblock Out-of-distribution detection in multi-label datasets using latent
  space of $\beta$-vae.
\newblock \emph{arXiv preprint arXiv:2003.08740}, 2020.

\bibitem[Tomczak \& Welling(2018)Tomczak and Welling]{tomczak2018vae}
Tomczak, J. and Welling, M.
\newblock Vae with a vampprior.
\newblock In \emph{International Conference on Artificial Intelligence and
  Statistics}, pp.\  1214--1223. PMLR, 2018.

\bibitem[Tsoumakas et~al.(2008)Tsoumakas, Katakis, and
  Vlahavas]{tsoumakas2008effective}
Tsoumakas, G., Katakis, I., and Vlahavas, I.
\newblock Effective and efficient multilabel classification in domains with
  large number of labels.
\newblock In \emph{Proc. ECML/PKDD 2008 Workshop on Mining Multidimensional
  Data (MMD’08)}, volume~21, pp.\  53--59, 2008.

\bibitem[Tu \& Gimpel(2018)Tu and Gimpel]{tu2018learning}
Tu, L. and Gimpel, K.
\newblock Learning approximate inference networks for structured prediction.
\newblock \emph{arXiv preprint arXiv:1803.03376}, 2018.

\bibitem[Wang et~al.(2021)Wang, Liu, Guo, and Sun]{wang2020unsupervised}
Wang, F., Liu, H., Guo, D., and Sun, F.
\newblock Unsupervised representation learning by invariancepropagation.
\newblock \emph{arXiv preprint arXiv:2010.11694}, 2021.

\bibitem[Wang et~al.(2014)Wang, Song, Leung, Rosenberg, Wang, Philbin, Chen,
  and Wu]{wang2014learning}
Wang, J., Song, Y., Leung, T., Rosenberg, C., Wang, J., Philbin, J., Chen, B.,
  and Wu, Y.
\newblock Learning fine-grained image similarity with deep ranking.
\newblock In \emph{Proceedings of the IEEE Conference on Computer Vision and
  Pattern Recognition}, pp.\  1386--1393, 2014.

\bibitem[Wang et~al.(2016)Wang, Yang, Mao, Huang, Huang, and Xu]{wang2016cnn}
Wang, J., Yang, Y., Mao, J., Huang, Z., Huang, C., and Xu, W.
\newblock Cnn-rnn: A unified framework for multi-label image classification.
\newblock In \emph{CVPR}, 2016.

\bibitem[Wang \& Wang(2019)Wang and Wang]{wang2019neural}
Wang, P.~Z. and Wang, W.~Y.
\newblock Neural gaussian copula for variational autoencoder.
\newblock \emph{arXiv preprint arXiv:1909.03569}, 2019.

\bibitem[Wu \& Goodman(2018)Wu and Goodman]{wu2018multimodal}
Wu, M. and Goodman, N.
\newblock Multimodal generative models for scalable weakly-supervised learning.
\newblock In \emph{Advances in Neural Information Processing Systems}, 2018.

\bibitem[Yeh et~al.(2017)Yeh, Wu, Ko, and Wang]{yeh2017learning}
Yeh, C.-K., Wu, W.-C., Ko, W.-J., and Wang, Y.-C.~F.
\newblock Learning deep latent space for multi-label classification.
\newblock In \emph{AAAI}, 2017.

\bibitem[Yu et~al.(2013)Yu, Rangwala, Domeniconi, Zhang, and Yu]{yu2013protein}
Yu, G., Rangwala, H., Domeniconi, C., Zhang, G., and Yu, Z.
\newblock Protein function prediction using multilabel ensemble classification.
\newblock \emph{IEEE/ACM Transactions on Computational Biology and
  Bioinformatics}, 2013.

\bibitem[Zhang \& Zhou(2007)Zhang and Zhou]{zhang2007ml}
Zhang, M.-L. and Zhou, Z.-H.
\newblock Ml-knn: A lazy learning approach to multi-label learning.
\newblock \emph{Pattern recognition}, 40\penalty0 (7):\penalty0 2038--2048,
  2007.

\bibitem[Zhang \& Zhou(2013)Zhang and Zhou]{zhang2013review}
Zhang, M.-L. and Zhou, Z.-H.
\newblock A review on multi-label learning algorithms.
\newblock \emph{IEEE transactions on knowledge and data engineering}, 2013.

\bibitem[Zhang et~al.(2018)Zhang, Li, Liu, and Geng]{zhang2018binary}
Zhang, M.-L., Li, Y.-K., Liu, X.-Y., and Geng, X.
\newblock Binary relevance for multi-label learning: an overview.
\newblock \emph{Frontiers of Computer Science}, 12\penalty0 (2):\penalty0
  191--202, 2018.

\end{thebibliography}
\bibliographystyle{icml2022}

\clearpage
\appendix

\section{Contrastive Learning Module} \label{appx:cl}

\subsection{Connection with Triplet Loss}

Triplet loss \cite{wang2014learning} is one of the popular ranking losses used in multi-label learning \cite{seymour2018multi}. 

Given an anchor embedding $v_x^f$, a positive embedding $v_+$ and a negative embedding $v_-$, they form a triplet $(v_x^f, v_+, v_-)$. A triplet loss is defined as 
\begin{equation}
\begin{aligned}
&\mathcal L_{trip}(v_x^f, v_+, v_-)\\
=&max\{0, g+dist(v_x^f, v_+)-dist(v_x^f,v_-)\}
\end{aligned}
\label{eq:trip}
\end{equation}
where $g$ is a gap parameter measuring the distance between $(v_x^f, v_+)$ and $(v_x^f, v_-)$, and $dist(\cdot,\cdot)$ is a distance function. This hinge loss $\mathcal L_{trip}$ encourages fewer violations to ``positive$>$negative" ranking order.
Let $\tau=1/2$. With the same triplet, we can write down a contrastive loss
\begin{equation}
\begin{aligned}
&\mathcal L_{CL}(v_x^f, v_+, v_-)\\
=&-\log \frac{\exp(2\cdot v^f_x\cdot v_+)}{\sum_{t\in \{+,-\}}\exp(2\cdot v^f_x\cdot v_t)}\\
=&\log (1+\frac{\exp(2\cdot v^f_x\cdot v_-)}{\exp(2\cdot v^f_x\cdot v_+)})\\
\approx&1+(2\cdot v^f_x\cdot v_--2\cdot v^f_x\cdot v_+)\\
=&1+(-v_x^f\cdot v_x^f+2v^f_x\cdot v_--v_-\cdot v_-\\
&+v_x^f\cdot v_x^f-2\cdot v^f_x\cdot v_++v_+\cdot v_+)\\
=&||v_x^f-v_+||^2+||v_x^f-v_-||^2+1
\end{aligned}
\label{eq:trip_cl}
\end{equation}
Note that in the second to the last equation, $v_+$ and $v_-$ have the same norm due to the normalization in our contrastive learning module. 

By setting $dist(\cdot,\cdot)$ to commonly used $\ell_2$ distance and $g=1$, Eq.~\ref{eq:trip_cl} is a fair approximation of Eq.~\ref{eq:trip}. Therefore, triplet loss can be viewed as a special case of contrastive loss. In contrastive loss, embeddings are normalized and more positives/negatives are available. As shown in \cite{chen2020simple}, contrastive loss generally outperforms triplet loss.

\subsection{Gradients of Contrastive Loss}

Recall our contrastive loss:
\begin{equation}
\begin{aligned}
\mathcal L_{CL}=\sum_{(x,y)\in \mathcal B}\frac{1}{|P(y)|}\sum_{p\in P(y)}-\log \frac{\exp(v^f_x\cdot v_p^l/\tau)}{\sum_{t\in A}\exp(v^f_x\cdot v_t^l/\tau)}
\end{aligned}
\end{equation}

For the illustration purpose, we only consider one sample $(x,y)$ instead of one batch:
\begin{equation}
\begin{aligned}
\mathcal L_{CL}=\frac{1}{|P(y)|}\sum_{p\in P(y)}-\log \frac{\exp(v^f_x\cdot v_p^l/\tau)}{\sum_{t\in A}\exp(v^f_x\cdot v_t^l/\tau)}
\end{aligned}
\end{equation}

Define $N(y)\equiv A\setminus P(y)$. We now derive the gradients w.r.t. $v_x^f$. 

\begin{equation}
\begin{aligned}
\frac{\partial\mathcal L_{CL}}{\partial v_x^f}=&\frac{1}{\tau|P(y)|}\sum_{p\in P(y)}(\frac{\sum_{t\in A}v_t^l\exp(v^f_x\cdot v_t^l/\tau)}{\sum_{t\in A}\exp(v^f_x\cdot v_t^l/\tau)}-v_p^l)\\
=&\frac{1}{\tau|P(y)|}\sum_{p\in P(y)}(\frac{\sum_{t\in P(y)}v_t^l\exp(v^f_x\cdot v_t^l/\tau)}{\sum_{t\in A}\exp(v^f_x\cdot v_t^l/\tau)}+\\
&\frac{\sum_{t\in N(y)}v_t^l\exp(v^f_x\cdot v_t^l/\tau)}{\sum_{t\in A}\exp(v^f_x\cdot v_t^l/\tau)}-v_p^l)\\
=&\frac{1}{\tau}\frac{\sum_{t\in P(y)}v_t^l\exp(v^f_x\cdot v_t^l/\tau)}{\sum_{t\in A}\exp(v^f_x\cdot v_t^l/\tau)}+\\
&\frac{1}{\tau}\frac{\sum_{t\in N(y)}v_t^l\exp(v^f_x\cdot v_t^l/\tau)}{\sum_{t\in A}\exp(v^f_x\cdot v_t^l/\tau)}-\\
&\frac{1}{\tau|P(y)|}\sum_{p\in P(y)}v_p^l\\
=&\frac{1}{\tau}[\sum_{t\in P(y)}v_t^l(\frac{\exp(v^f_x\cdot v_t^l/\tau)}{\sum_{a\in A}\exp(v^f_x\cdot v_a^l/\tau)}-\frac{1}{|P(y)|})+\\
&\sum_{t\in N(y)}v_t^l\frac{\exp(v^f_x\cdot v_t^l/\tau)}{\sum_{a\in A}\exp(v^f_x\cdot v_a^l/\tau)}]\\
\end{aligned}
\end{equation}

Further, we have the unnormalized feature embedding $w_x^f$, $v_x^f=\frac{w_x^f}{||w_x^f||}$.
\begin{equation}
\begin{aligned}
\frac{\partial v_x^f}{\partial w_x^f}=&\frac{1}{||w_x^f||}(I-\frac{w_x^f{w_x^f}^T}{||w_x^f||^2})=\frac{1}{||w_x^f||}(I-v_x^f{v_x^f}^T)
\end{aligned}    
\end{equation}
where $I$ is an $E\times E$ identity matrix. 
The gradient of $\mathcal L_{CL}$ w.r.t. $w_x^f$ can be derived with chain rule,
\begin{equation}
\begin{aligned}
&\frac{\partial \mathcal L_{CL}}{\partial w_x^f}
=\frac{\partial v_x^f}{\partial w_x^f}\frac{\partial\mathcal L_{CL}}{\partial v_x^f}\\
=&\frac{1}{||w_x^f||}(I-v_x^f{v_x^f}^T)\frac{1}{\tau}[\sum_{t\in P(y)}v_t^l(\frac{\exp(v^f_x\cdot v_t^l/\tau)}{\sum_{a\in A}\exp(v^f_x\cdot v_a^l/\tau)}\\
&-\frac{1}{|P(y)|})+\sum_{t\in N(y)}v_t^l\frac{\exp(v^f_x\cdot v_t^l/\tau)}{\sum_{a\in A}\exp(v^f_x\cdot v_a^l/\tau)}]\\
=&\frac{1}{\tau||w_x^f||}[\sum_{t\in P(y)}(v_t^l-(v_x^fv_t^l)v_x^f)(\frac{\exp(v^f_x\cdot v_t^l/\tau)}{\sum_{a\in A}\exp(v^f_x\cdot v_a^l/\tau)}\\
&-\frac{1}{|P(y)|})+\\
&\sum_{t\in N(y)}(v_t^l-(v_x^fv_t^l)v_x^f)\frac{\exp(v^f_x\cdot v_t^l/\tau)}{\sum_{a\in A}\exp(v^f_x\cdot v_a^l/\tau)}]\\
\end{aligned}    
\end{equation}

We can then observe that if $v_x^f$ and $v_t^l$ are orthogonal ($v_x^fv_t^l\to 0$), $||v_t^l-(v_x^fv_t^l)v_x^f||$ will be close to 1 and the gradients would be large. Otherwise, for weak positives or negatives ($|v_x^fv_t^l|\to 1$), the gradients would be small.

\begin{figure}[t]
\centering
\includegraphics[width=\linewidth]{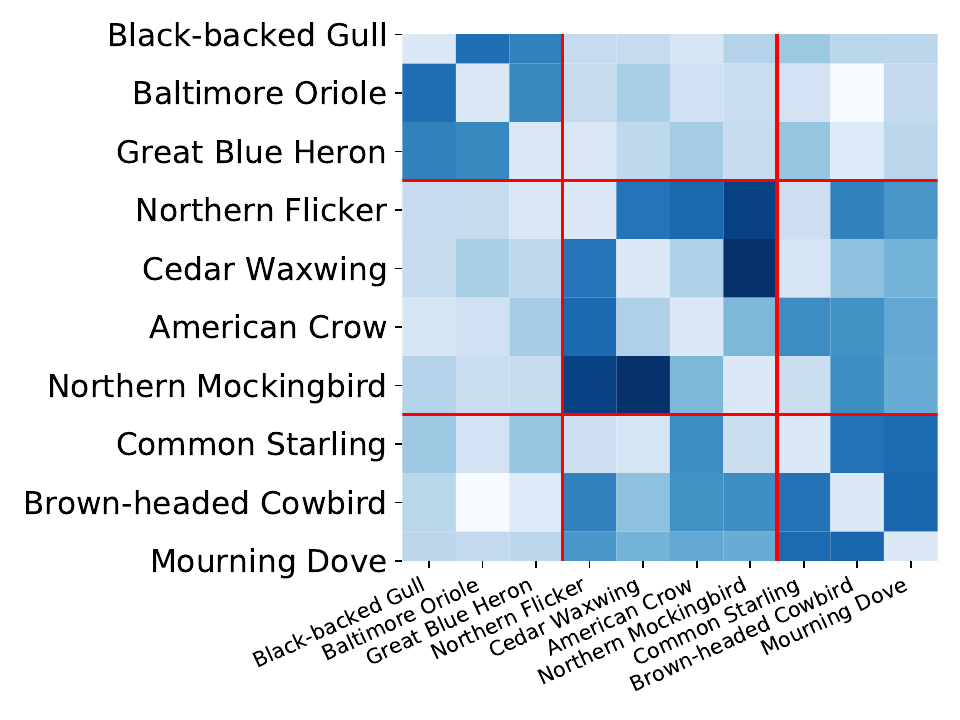}
\caption{Label-label inner-products by MPVAE.}
\label{fig:bird_mpvae}
\end{figure}

\begin{figure}
\centering
\begin{minipage}{0.8\linewidth}
  \centering
  \includegraphics[width=\linewidth]{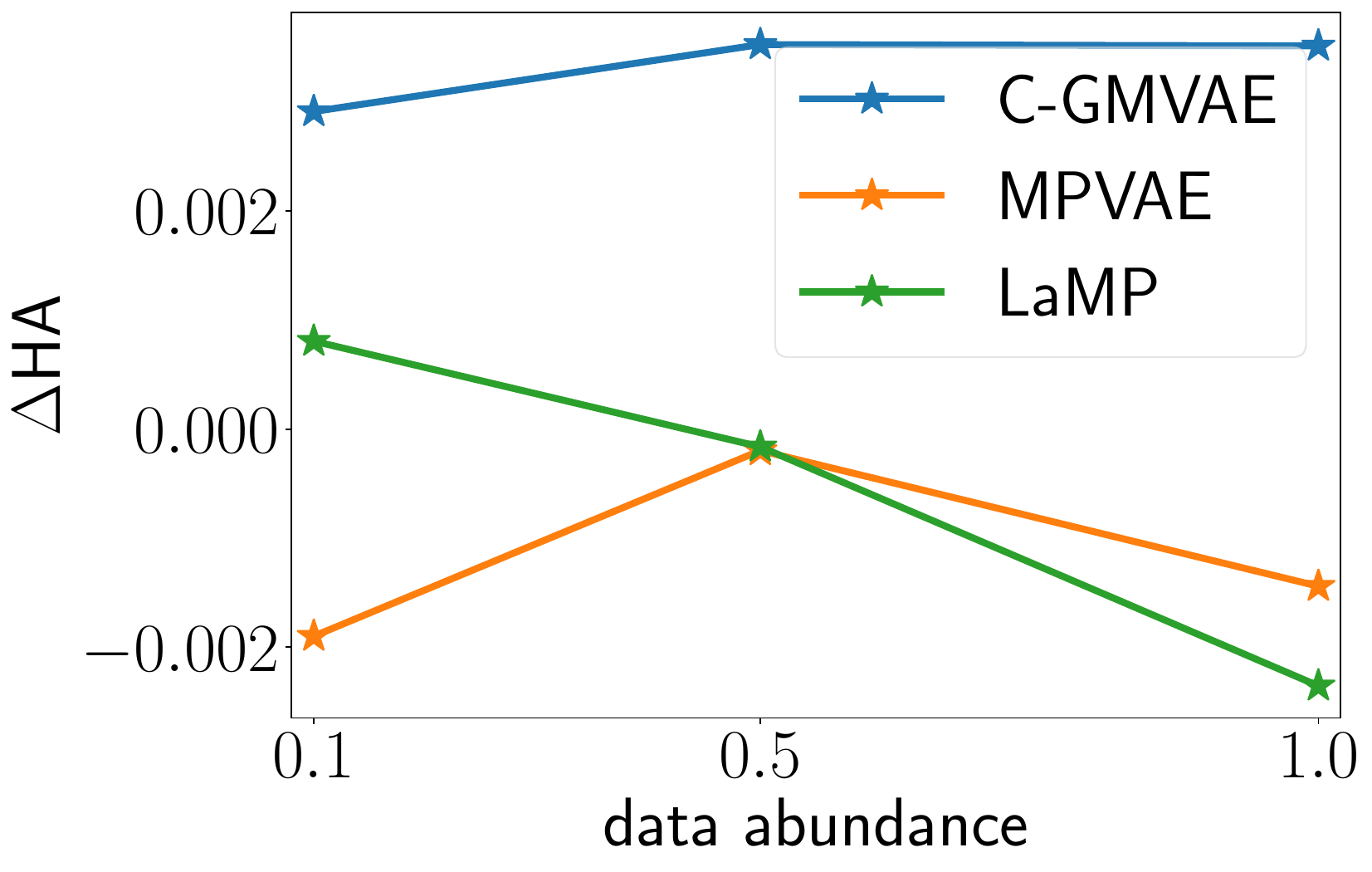}
\end{minipage}~~~~~
~\\
\begin{minipage}{0.75\linewidth}
  \centering
  \includegraphics[width=\linewidth]{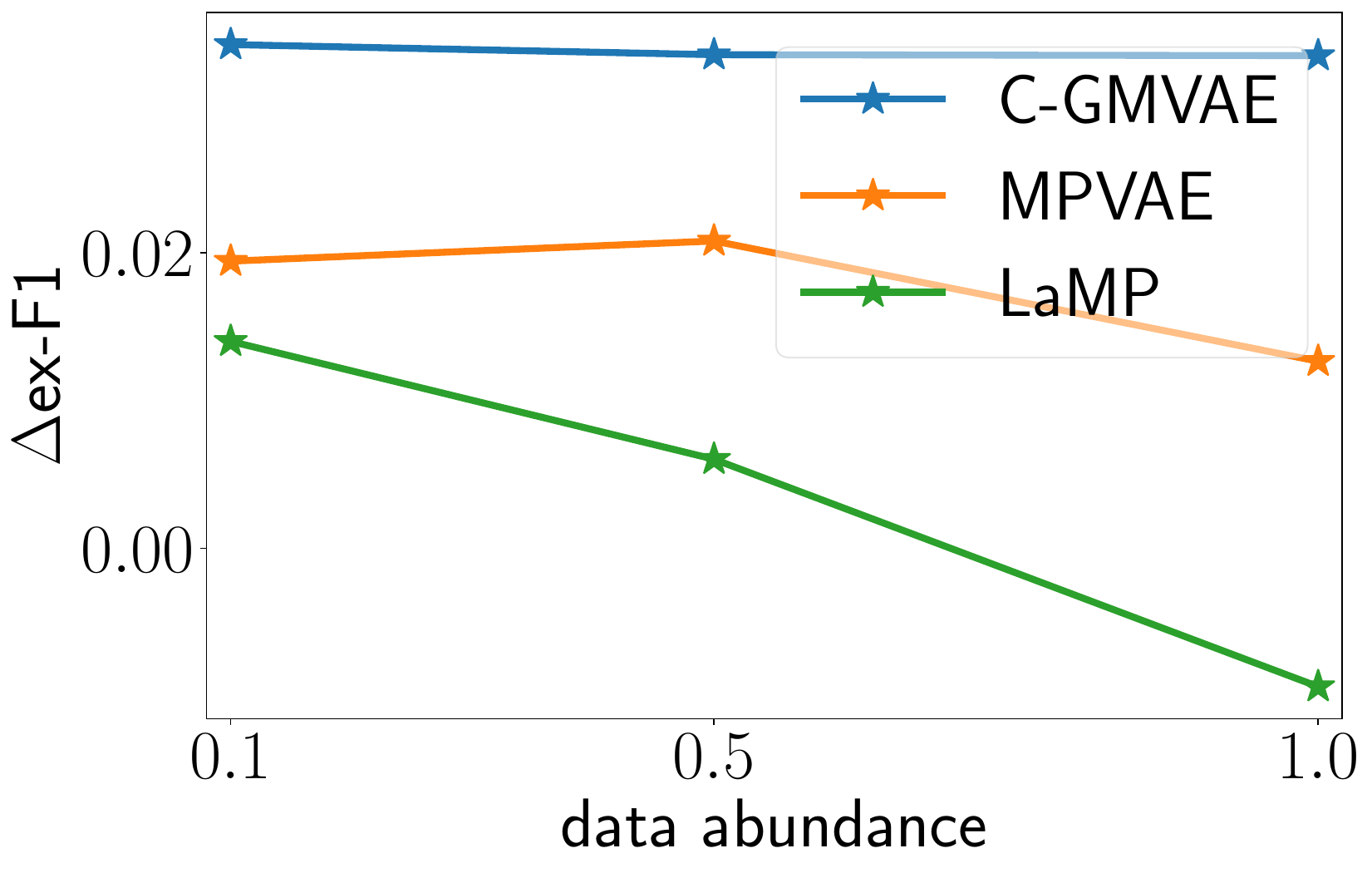}
\end{minipage}
~\\
\begin{minipage}{0.75\linewidth}
  \centering
  \includegraphics[width=\linewidth]{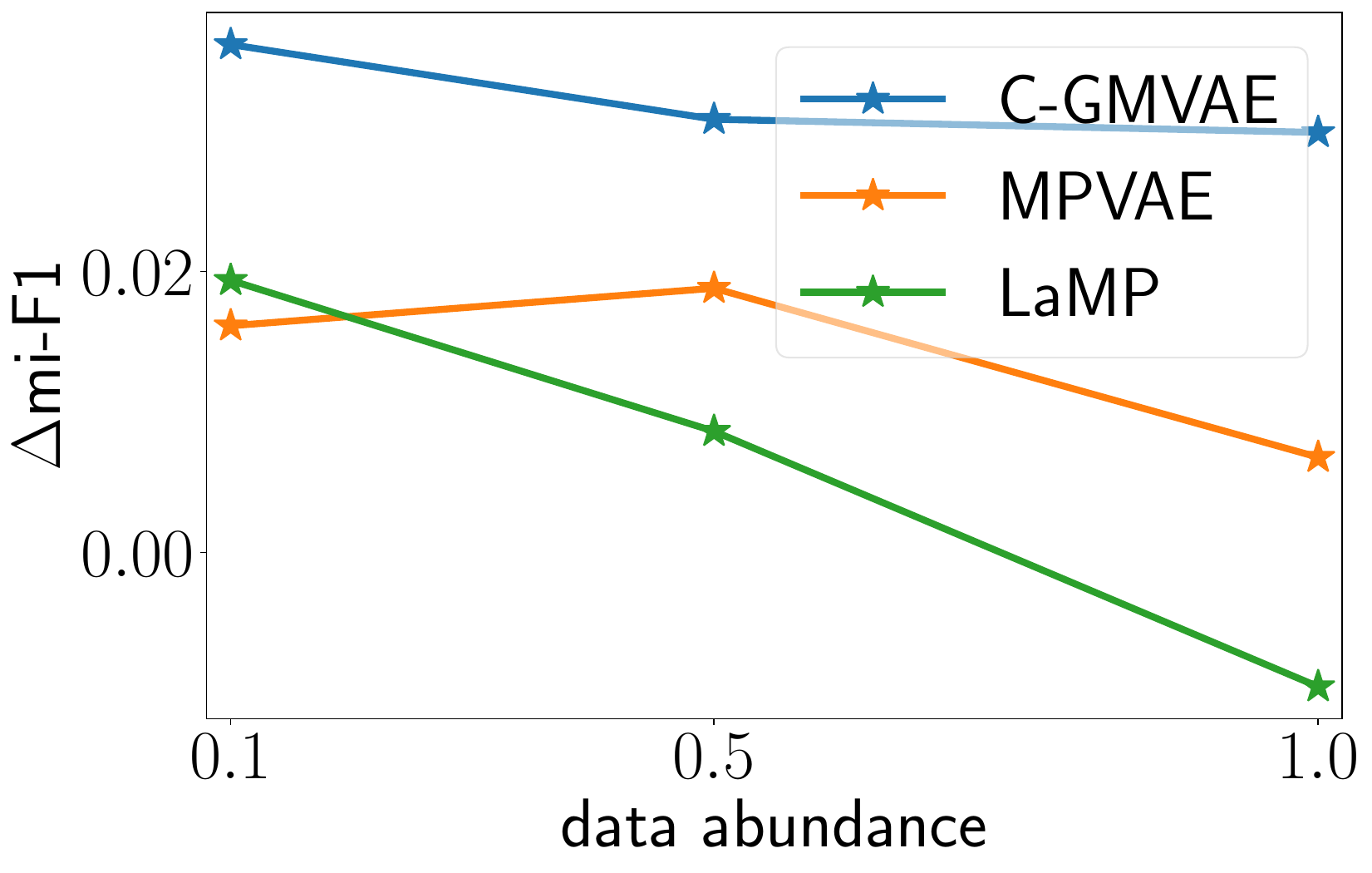}
\end{minipage}
~\\
\begin{minipage}{0.75\linewidth}
  \centering
  \includegraphics[width=\linewidth]{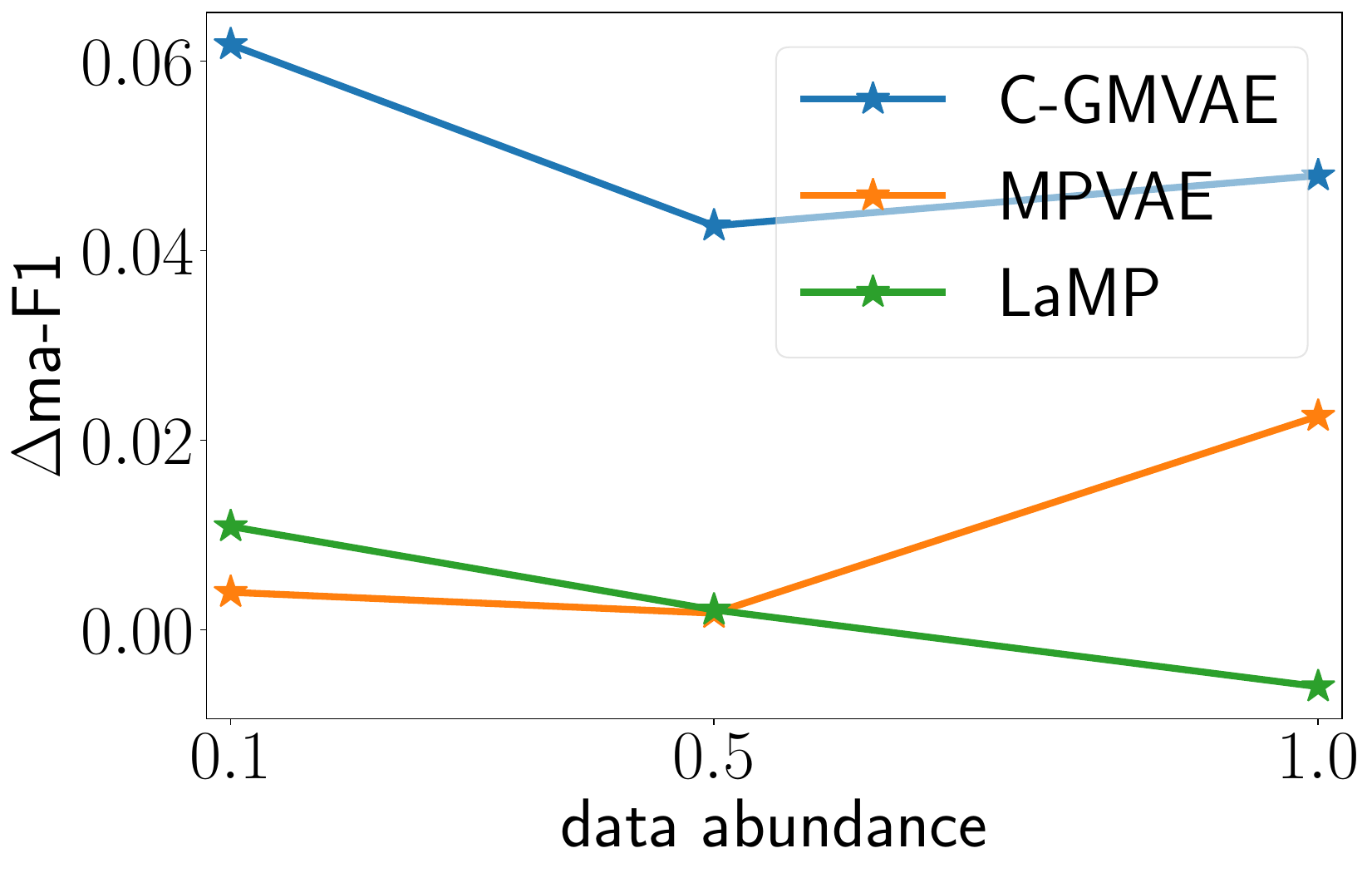}
\end{minipage}
\caption{Relative performances w.r.t. HA, ex-F1, mi-F1 and ma-F1 on \textit{mirflickr} dataset.}
\label{fig:semi_mirflickr}
\end{figure}

\begin{figure}
\centering
\begin{minipage}{0.8\linewidth}
  \centering
  \includegraphics[width=\linewidth]{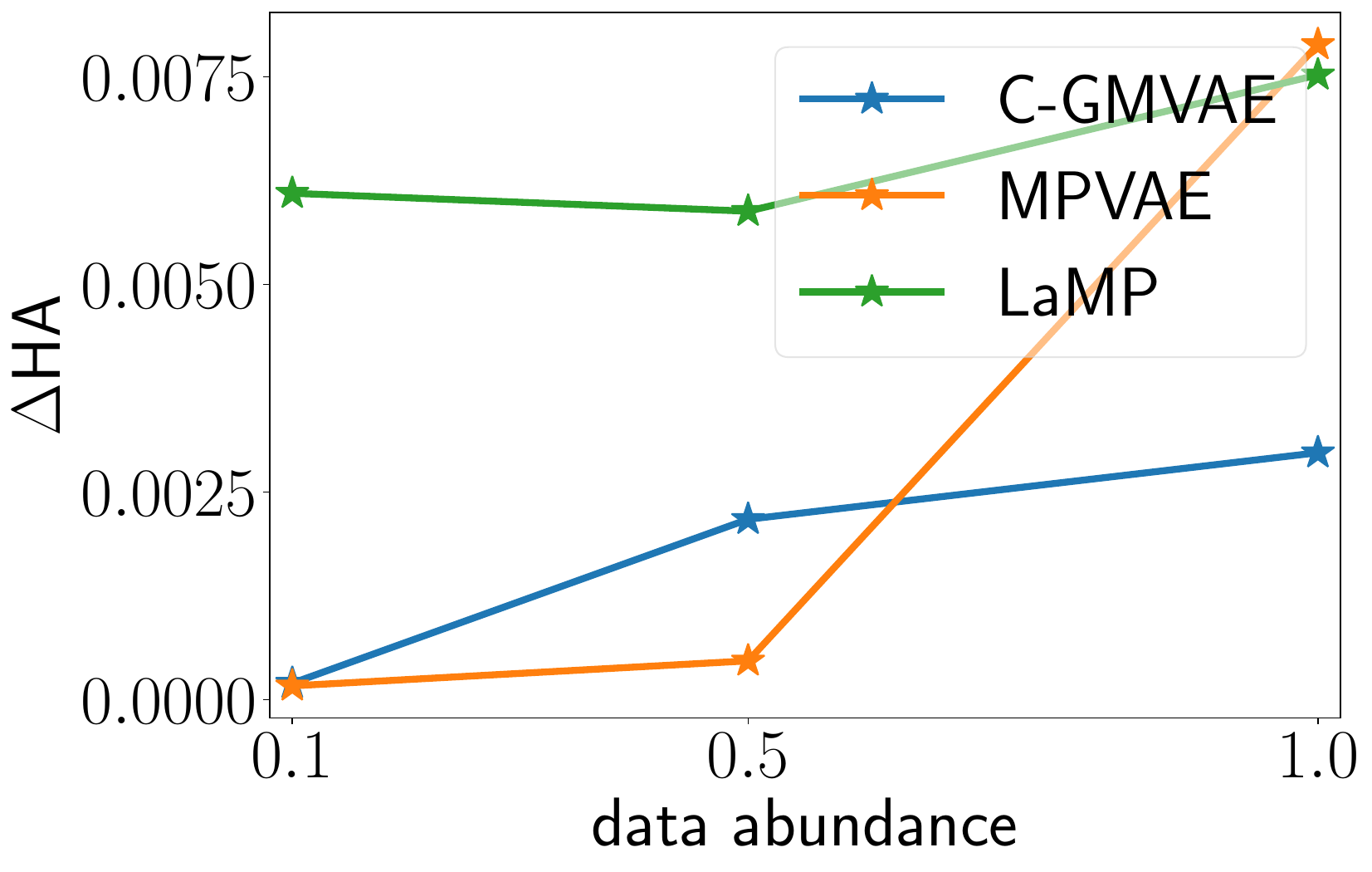}
\end{minipage}~~~~~
~\\
\begin{minipage}{0.79\linewidth}
  \centering
  \includegraphics[width=\linewidth]{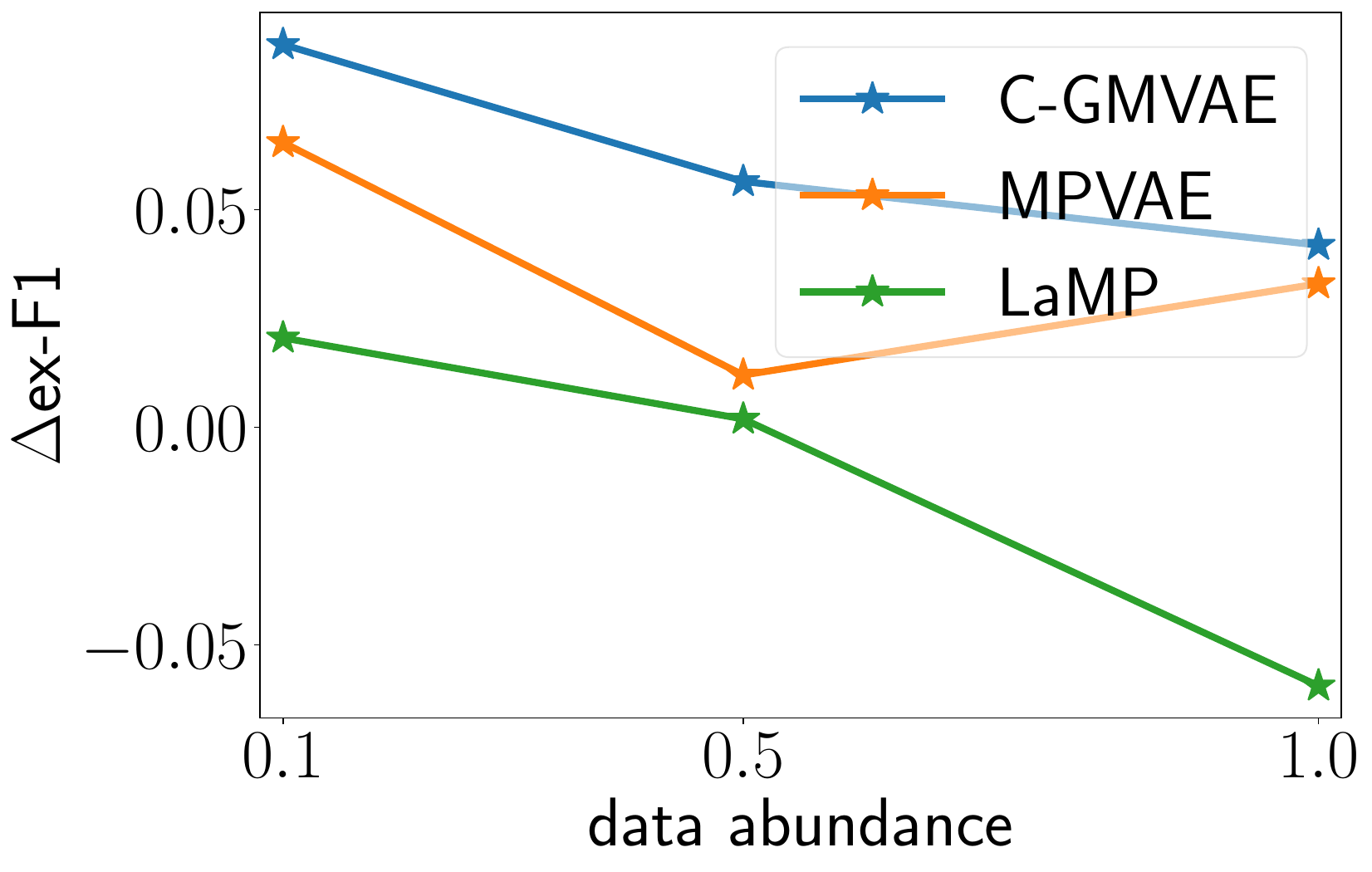}
\end{minipage}~~~~
~\\
\begin{minipage}{0.75\linewidth}
  \centering
  \includegraphics[width=\linewidth]{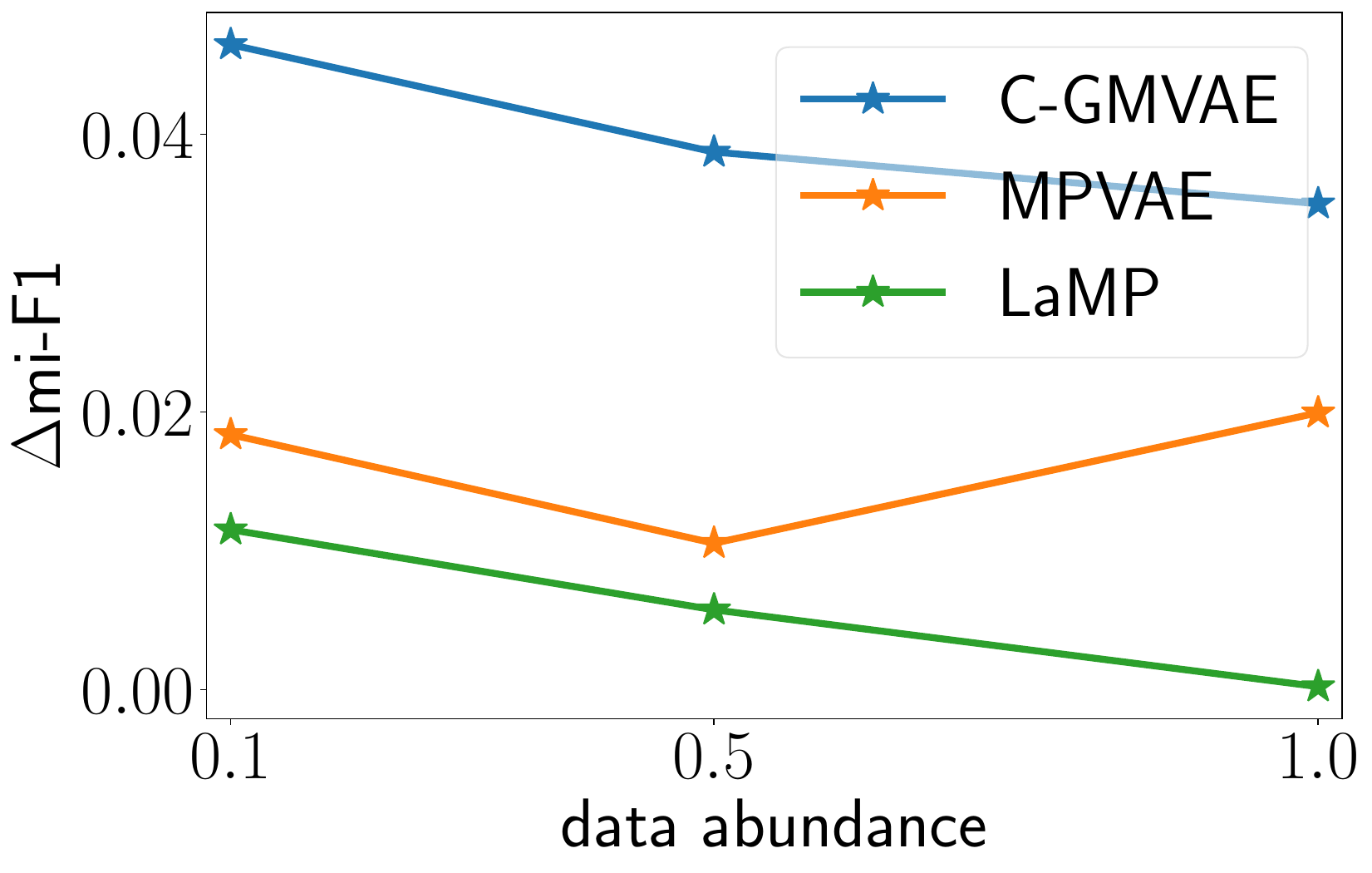}
\end{minipage}
~\\
\begin{minipage}{0.75\linewidth}
  \centering
  \includegraphics[width=\linewidth]{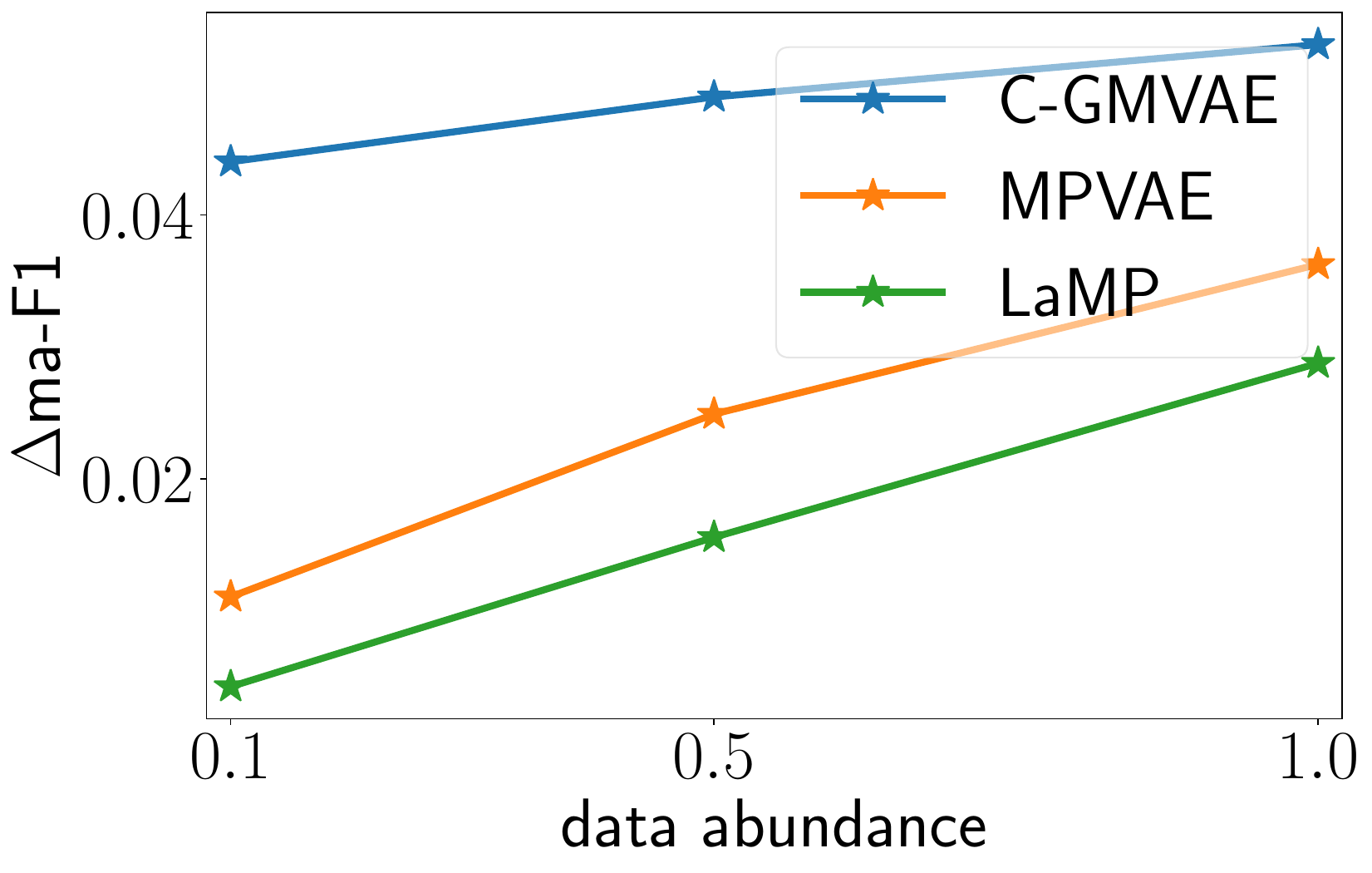}
\end{minipage}
\caption{Relative performances w.r.t. HA, ex-F1, mi-F1 and ma-F1 on \textit{nus-vec} dataset.}
\label{fig:semi_nusvec}
\end{figure}

\begin{figure}
\centering
\begin{minipage}{0.75\linewidth}
  \centering
  \includegraphics[width=\linewidth]{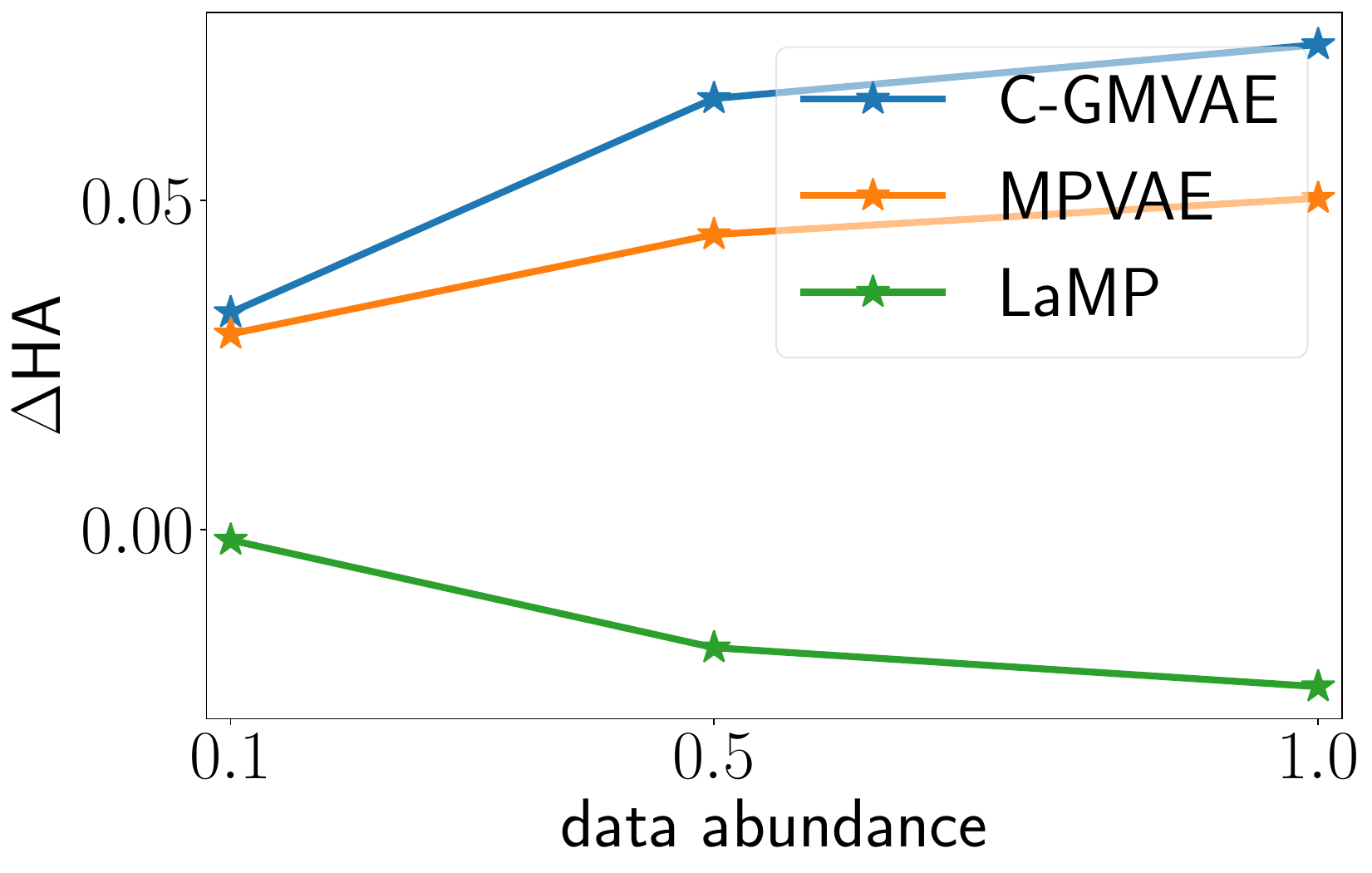}
\end{minipage}
~\\
\begin{minipage}{0.75\linewidth}
  \centering
  \includegraphics[width=\linewidth]{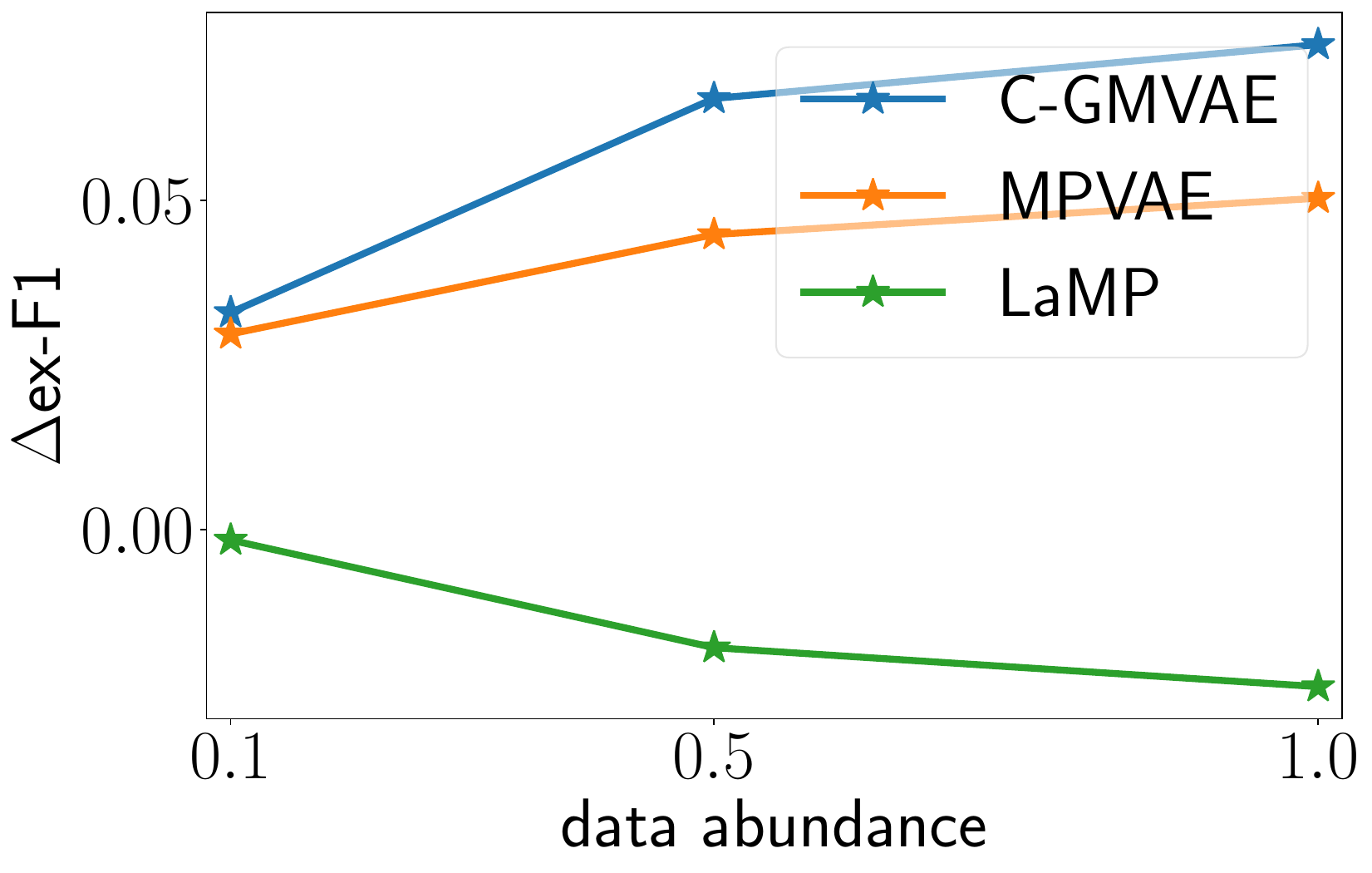}
\end{minipage}
~\\
\begin{minipage}{0.75\linewidth}
  \centering
  \includegraphics[width=\linewidth]{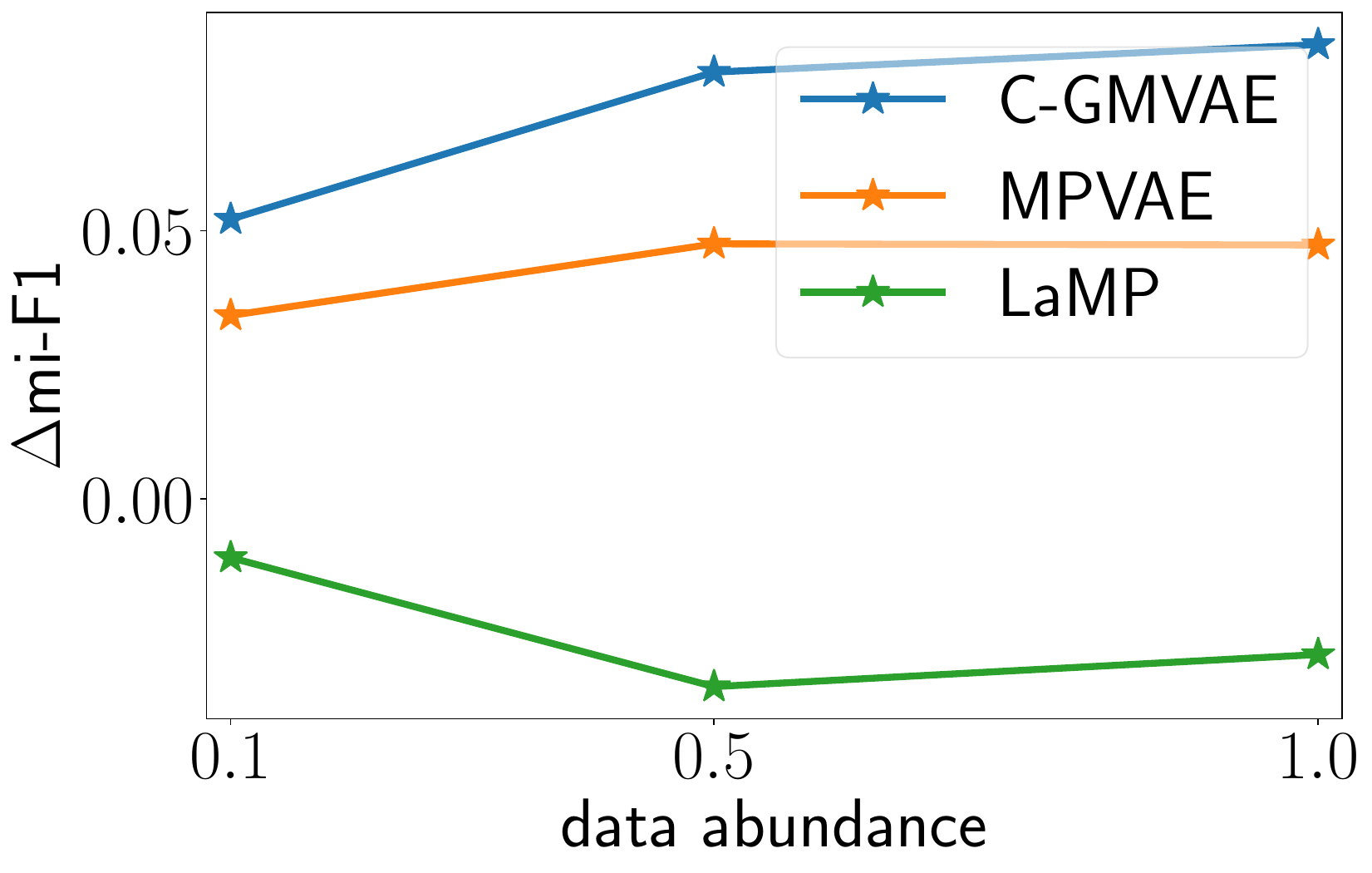}
\end{minipage}
~\\
\begin{minipage}{0.78\linewidth}
  \centering
  \includegraphics[width=\linewidth]{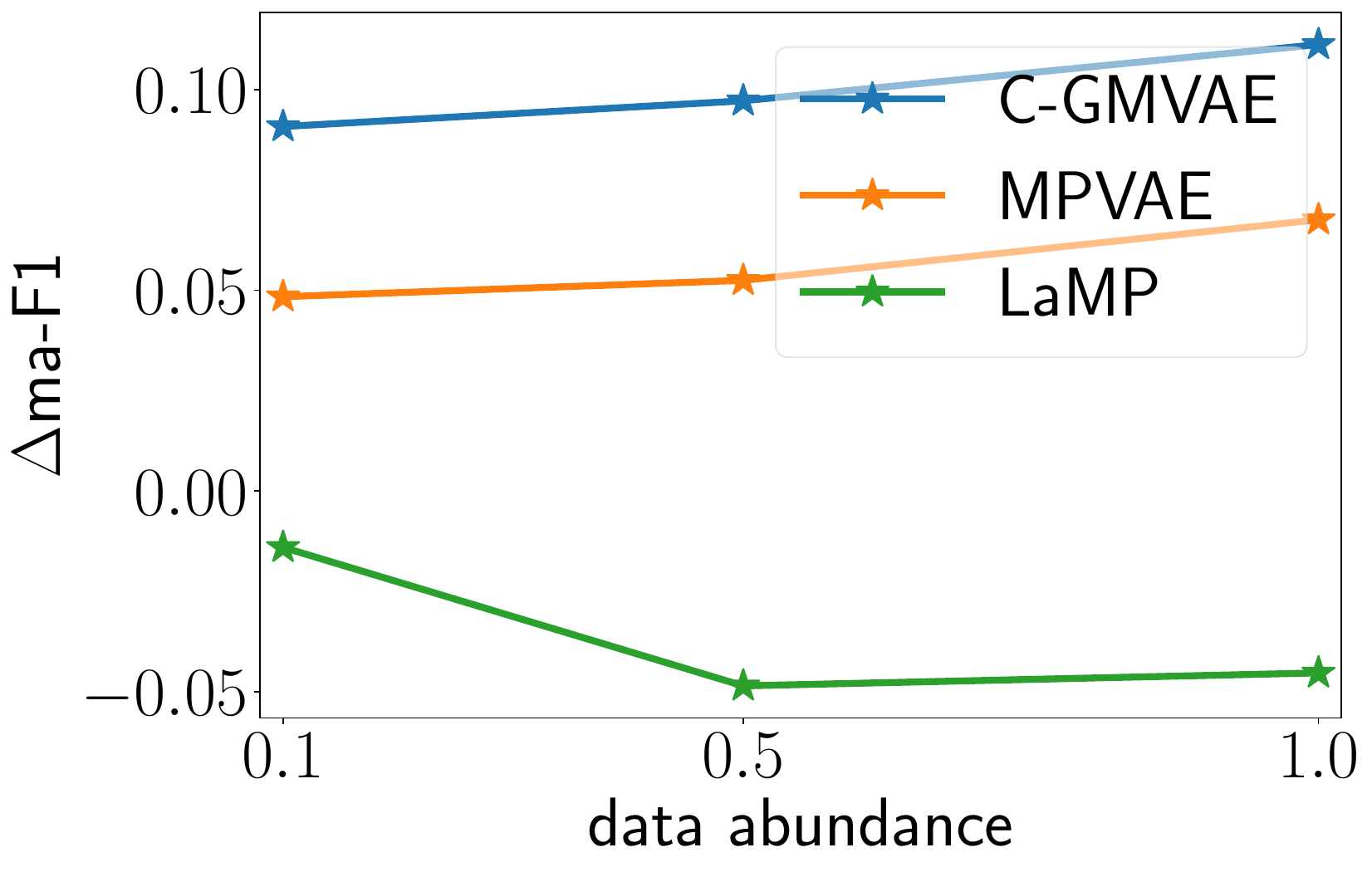}
\end{minipage}~~~
\caption{Relative performances w.r.t. HA, ex-F1, mi-F1 and ma-F1 on \textit{ebird} dataset.}
\label{fig:semi_ebird}
\end{figure}

\section{Supplementary Experimental Results}




\subsection{Implementation Details}
We use one Tesla V100 GPU on CentOS for every experiment. The batch size is set to 128. The latent dimensionality is 64. The feature encoder is an MLP with 3 hidden layers of sizes [256, 512, 256]. The label encoder has 2 hidden layers of sizes [512, 256]. The decoder contains 2 hidden layers of sizes [512, 512]. On \textit{reuters} and \textit{bookmarks}, we add one more hidden layer with 512 units to the decoder. The embedding size $E$ is 2048 (tuned within the range [512, 1024, 2048, 3072]). We set $\alpha=1$ (tuned within [0.1, 0.5, 1, 1.5, 2]), $\beta=0.5$ (tuned within [0.1, 0.5, 1, 1.5, 2.0]) for most runs. We tune learning rates from 0.0001 to 0.004 with interval 0.0002, dropout ratio from [0.3, 0.5, 0.7], and weight decay from [0, 0.01, 0.0001]. Grid search is adopted for tuning. 
Every batch in our experiments requires less than 16GB memory. The number of epochs is 100 by default. 

\begin{table}[h]
\centering
\scalebox{0.72}{
\begin{tabular}{c|cc|cc}
\toprule
\multirow{2}{*}{model} & \multicolumn{2}{c|}{eBird (s)} & \multicolumn{2}{c}{mirflickr (s)} \\ \cline{2-5} 
 & \multicolumn{1}{c|}{Train (per epoch)} & Test (total) & \multicolumn{1}{c|}{Train (per epoch)} & Test (total) \\ \hline
C2AE & \multicolumn{1}{c|}{27} & 29 & \multicolumn{1}{c|}{18} & 24 \\ \hline
LaMP & \multicolumn{1}{c|}{32} & 33 & \multicolumn{1}{c|}{22} & 31 \\ \hline
MPVAE & \multicolumn{1}{c|}{43} & 97 & \multicolumn{1}{c|}{25} & 57 \\ \hline
ASL & \multicolumn{1}{c|}{24} & 25 & \multicolumn{1}{c|}{20} & 26 \\ \hline
RBCC & \multicolumn{1}{c|}{3840} & 275 & \multicolumn{1}{c|}{1320} & 337 \\ \hline
\midrule
C-GMVAE & \multicolumn{1}{c|}{22} & 24 & \multicolumn{1}{c|}{14} & 16 \\
\bottomrule
\end{tabular}
}
\caption{Comparison of different models' time costs.}
\label{tab:time}
\end{table}

In Tab.~\ref{tab:time}, we show the per-epoch runtime for training, and the total time cost for testing. Our C-GMVAE is very competitive w.r.t. both training and testing time costs.





\subsection{Training on Fewer Data}

We provide relative performances of several major state-of-the-art methods including ours to C2AE, on HA, ex-F1, mi-F1, ma-F1 scores. All methods are trained on 10\% or 50\% of the data, including C2AE. The compared results have the same amount of data for training and thus the comparison is fair.

Fig.~\ref{fig:semi_mirflickr}, Fig.~\ref{fig:semi_nusvec}, Fig.~\ref{fig:semi_ebird} show the relative performance of various state-of-the-art methods over C2AE, on \textit{mirflickr}, \textit{nus-vec}, \textit{eBird} respectively.

\end{document}